\documentclass[lettersize,journal]{IEEEtran}
\usepackage{amsmath,amsfonts}
\usepackage{algorithmic}
\usepackage{algorithm}
\usepackage{array}
\usepackage[caption=false,font=normalsize,labelfont=sf,textfont=sf]{subfig}
\usepackage{textcomp}
\usepackage{stfloats}
\usepackage{url}
\usepackage{verbatim}
\usepackage{graphicx}
\usepackage{cite}
\usepackage{colortbl}
\usepackage{xcolor}
\definecolor{lightergray}{gray}{0.95}
\usepackage{multirow}
\usepackage{booktabs}
\usepackage{diagbox}
\usepackage{pifont}
\usepackage{makecell}
\usepackage{threeparttable}

\begin{document}

\title{PalmBridge: A Plug-and-Play Feature Alignment Framework for Open-Set Palmprint Verification}

\author{Chenke Zhang, Ziyuan Yang, Licheng Yan, Shuyi Li, Andrew Beng Jin Teoh \IEEEmembership{Senior Member, IEEE}, Bob Zhang \IEEEmembership{Senior Member, IEEE}, Yi Zhang \IEEEmembership{Senior Member, IEEE}
\thanks{This work has been submitted to the IEEE for possible publication. Copyright may be transferred without notice, after which this version may no longer be accessible.}
\thanks{C. Zhang, Z. Yang, and Y. Zhang are with the School of Cyber Science and Engineering, Sichuan University, Chengdu 610065, China (e-mail: zhangchankle@gmail.com, cziyuanyang@gmail.com, yzhang@scu.edu.cn).}
\thanks{L. Yan and B. Zhang  are with the Pattern Analysis and Machine Intelligence Group, Department of Computer and Information Science, University of Macau, Taipa, Macau, China (e-mail: yc48110@um.edu.mo, bobzhang@um.edu.mo).}
\thanks{S. Li is with the School of Information Science and Technology, Beijing University of Technology, Beijing 100124, China (e-mail: syli2022@bjut.edu.cn). }
\thanks{A.B.J. Teoh is with the School of Electrical and Electronic Engineering, College of Engineering, Yonsei University, Seoul, Republic of Korea (e-mail: bjteoh@yonsei.ac.kr).}
}

\markboth{Journal of \LaTeX\ Class Files,~Vol.~14, No.~8, August~2021}%
{Shell \MakeLowercase{\textit{et al.}}: A Sample Article Using IEEEtran.cls for IEEE Journals}


\maketitle

\begin{abstract}
Palmprint recognition is widely used in biometric systems, yet real-world performance often degrades due to feature distribution shifts caused by heterogeneous deployment conditions. Most deep palmprint models assume a closed and stationary distribution, leading to overfitting to dataset-specific textures rather than learning domain-invariant representations. Although data augmentation is commonly used to mitigate this issue, it assumes augmented samples can approximate the target deployment distribution, an assumption that often fails under significant domain mismatch. To address this limitation, we propose PalmBridge, a plug-and-play feature-space alignment framework for open-set palmprint verification based on vector quantization. Rather than relying solely on data-level augmentation, PalmBridge learns a compact set of representative vectors directly from training features. During enrollment and verification, each feature vector is mapped to its nearest representative vector under a minimum-distance criterion, and the mapped vector is then blended with the original vector. This design suppresses nuisance variation induced by domain shifts while retaining discriminative identity cues. The representative vectors are jointly optimized with the backbone network using task supervision, a feature-consistency objective, and an orthogonality regularization term to form a stable and well-structured shared embedding space. Furthermore, we analyze feature-to-representative mappings via assignment consistency and collision rate to assess model's sensitivity to blending weights. Experiments on multiple palmprint datasets and backbone architectures show that PalmBridge consistently reduces EER in intra-dataset open-set evaluation and improves cross-dataset generalization with negligible to modest runtime overhead.\footnote{Code will be made publicly available at https://github.com/Chankle-Z/PalmBridge.}

\end{abstract}

\begin{IEEEkeywords}
Palmprint verification, vector quantization, open-set verification, biometric recognition
\end{IEEEkeywords}

\IEEEpeerreviewmaketitle

\section{Introduction}

\IEEEPARstart{P}{almprint} verification has attracted widespread attention and extensive practical applications due to the high individuality and richness of palmprint patterns. Palmprint images contain abundant and distinctive features, including principal lines, wrinkles, ridges, and fine-grained texture~\cite{zhang2003online}, which are difficult to forge and thus provide high reliability for identity verification. Besides, palmprint verification is inherently user-friendly and health-conscious, as it typically relies on non-invasive and contactless acquisition processes~\cite{gao2025deep}.

With increasing research interest in palmprint verification, various methods have been developed. In particular, the introduction of deep learning (DL) techniques has significantly advanced this field, leading to highly effective and competitive verification performance. Consequently, palmprint verification systems have been successfully deployed in various practical scenarios, such as palmprint-based payment systems that enhance convenience and usability in daily life.

Despite these advances, most existing methods are developed and evaluated under relatively controlled conditions, where training and testing data are assumed to follow similar distributions. Under this assumption, these methods achieve promising performance; however, they often struggle to generalize to unseen individuals captured by the same device or to previously enrolled identities acquired under different environmental conditions. Such limitations lead to notable performance degradation in practical deployments involving new users or changing acquisition environments. A key challenge stems from task-irrelevant variations in palmprint images, such as changes in hand posture, illumination, region-of-interest~(ROI) localization errors, sensor noise, and acquisition conditions, which induce feature-distribution shifts and consequently impair recognition accuracy.

To address this challenge, numerous methods have recently been proposed 
to improve the robustness and generalization ability of palmprint verification models.
Among them, domain generalization (DG)-based methods \cite{zhou2022domain} aim to learn domain-invariant representations from multiple source domains, which enables models to generalize to unseen target domains without requiring additional adaptation. 
However, most DG methods place heavy emphasis on the training data, either relying on high-quality multi-domain samples or employing extensive data augmentation to artificially enlarge the dataset. For example, Shen~\textit{et al.}~\cite{shen2022distribution} put forward a novel distribution-based loss to alleviate heterogeneity arising from variations in image characteristics. Their approach aims to enhance performance by enlarging the margin between intra-class and inter-class distributions. Specifically, the loss function comprises a positive histogram loss to pull positive pairs closer, a negative histogram loss to push negative pairs apart, and an ArcFace loss \cite{deng2019arcface} for identity classification. Nevertheless, constructing positive pairs requires palmprint images collected from different datasets acquired from distinct devices, which is often impractical in real-world conditions. 

More recently, Shao~\textit{et al.} \cite{shao2024learning} proposed a Palmprint Data and Feature Generation~(PDFG) method to improve model generalization to unseen datasets. This method exchanges palmprint feature information across different datasets during training to enhance cross-domain robustness. In contrast, Jia \textit{et al.} \cite{jia2025single} introduced PalmRSS, a single-source domain generalization method that reduces the dependency on multi-domain databases. PalmRSS integrates a Fourier alignment transform with histogram matching to minimize distribution differences across subsets, thereby improving generalization capability under limited training data.

Among existing approaches, particular attention has been given to the design and manipulation of training data. Most approaches attempt to enhance model generalization by performing feature exchange or mixing within the training set, with the goal of simulating domain heterogeneity and enlarging the effective training distribution. While such strategies can improve robustness to domain shifts to some extent, they remain fundamentally confined to the seen training domains and cannot explicitly handle unseen samples encountered in open-set verification scenarios. Besides, their generalization capability is limited to the simulated heterogeneity, which often fails to faithfully capture the variability present in real-world, unseen images. Moreover, feature exchange among training samples may stretch or even overlap the feature distributions of the same class, potentially weakening class separability. 

In palmprint open-set verification, the term \textit{``open-set"} denotes that identities encountered during testing are entirely absent from the training set. The core challenge, therefore, lies in whether the embedding geometry learned from the training identities can effectively generalize to novel identities. In practice, verification performance often degrades because the learned embeddings become overly sensitive to task-irrelevant factors. Such sensitivity amplifies intra-class variation for unseen identities, reduces the similarity between genuine samples, and ultimately leads to ambiguous decision boundaries. 

From this perspective, the key challenge of open-set palmprint verification lies in the excessive sensitivity of feature representations to nuisance (verification-irrelevant) variations that are not explicitly modeled during training. This sensitivity causes unstable embeddings for unseen identities and motivates the need for a mechanism that can guide unseen identities toward stable and representative regions of the feature space learned from training data.

To address this challenge, we propose a novel verification framework specifically designed for unseen palmprint images. Specifically, we introduce \textbf{\textit{PalmBridge}}, a dedicated feature-space module that stores a compact set of quantized feature vectors extracted from the training dataset for subsequent matching and verification. PalmBridge is trained jointly with the palmprint model and imposes minimal constraints on the training dataset. Once trained, it can be seamlessly integrated into the verification pipeline and is compatible with various palmprint architectures. To maximize the effectiveness of the stored quantized vectors, we further impose an orthogonality constraint to encourage diversity among them, thereby preventing redundancy and excessive similarity within the representative set. 

PalmBridge uses a nearest-neighbor vector quantization~(VQ) operator that is superficially similar to the codebook assignment in VQ-based representation learning \cite{van2017neural}. However, PalmBridge is fundamentally different from VQ-based models. It functions as a deterministic feature-space alignment operator trained under a discriminative backbone, augmented with a feature-consistency loss and an orthogonality regularizer on the representative vectors. During inference, PalmBridge does not replace the original vector with a discrete codeword; instead, it blends the original and mapped vectors using controllable coefficients, explicitly exposing a tradeoff between nuisance suppression and identity preservation. Moreover, PalmBridge is applied symmetrically to both enrollment and query vectors to pull them toward a shared embedding space prior to similarity scoring. Its behavior can be interpreted as suppressing nuisance-induced variance by projecting features toward high-density regions of the training feature distribution.

Overall, our framework directly aligns features of unseen samples by pulling them toward a unified and structured feature domain, which enables a dynamic adaptation mechanism that more effectively addresses open-set palmprint verification. Unlike augmentation-centric training strategies, PalmBridge operates solely at the feature level and can be deployed with or without data augmentation. By mapping original vectors toward a controlled set of representative feature vectors during both enrollment and verification, PalmBridge regularizes the feature distribution and substantially improves robustness under deployment-induced heterogeneity.
The main contributions of this paper are summarized as follows:
\begin{enumerate}
    \item We propose PalmpBridge, a plug-and-play feature-space alignment framework for open-set palmprint verification. PalmBridge learns a compact, orthogonality-regularized codebook from training features and applies a symmetric nearest-vector mapping to both enrollment and query vectors, followed by a controllable blending with the original vectors to balance nuisance suppression against identity preservation.
    \item We introduce a feature-consistency objective together with an orthogonality regularizer to optimize the representative vectors, which encourages a stable, diverse, and well-structured shared embedding space while preserving discriminative identity information.
    \item We conduct extensive experiments under both intra-dataset and cross-dataset open-set protocols. The results demonstrate consistent verification gains across diverse backbone architectures, with negligible to modest computational overhead.
\end{enumerate}

\section{Related Works}
\subsection{Palmprint Recognition Methods for General Scenario}
Palmprint recognition technology has been applied across diverse practical scenarios, and numerous sophisticated approaches have been developed over the past decades. Generally, these approaches can be broadly divided into four categories: subspace-based, statistical, coding-based, and deep learning-based methods \cite{zhong2019decade}. Subspace-based techniques map palmprint images or their feature representations into a lower-dimensional space to enhance discriminative capability. Statistical approaches typically rely on hand-crafted feature extraction followed by statistical modeling or matching strategies for identity verification \cite{zhang2018combining}. Moreover, recent studies have demonstrated that combining heterogeneous features can further improve recognition performance~\cite{fei2021jointly}.

Coding-based methods aim to extract distinct texture features from palmprint images and encode them into compact representations for efficient matching. Early work by Zhang \textit{et al.} \cite{zhang2003online} proposed PalmCode, which employed a 2D Gabor filter to extract features that are subsequently coded for fast matching. Inspired by PalmCode, Kong \textit{et al.} \cite{kong2004competitive} introduced Competitive Code~(CompCode), which was the first to incorporate a competitive mechanism into palmprint recognition. With the continued development of coding-based methods, Gabor filters have been widely used due to their strong capability in texture representation. For example, to capture texture characteristics from multiple orientations, the Binary Orientation Co-occurrence Vector (BOCV) \cite{guo2009palmprint} employs six Gabor filters. The magnitude responses obtained from these filters are then transformed into a binary representation to construct the final template. Yang \textit{et al.} \cite{yang2023multi} designed the 2nd-order Texture Co-occurrence Code (2TCC) to leverage higher-order texture relationships for palmprint analysis, and further extended it to the Multiple-order Texture Co-occurrence Code~(MTCC), which combines texture cues of different orders to yield more robust feature representations.

Inspired by the success of deep learning in vision tasks, a growing body of research has explored DL-based approaches to palmprint recognition. As an example, Chai \textit{et al.}~\cite{chai2019boosting} utilized a convolutional neural network (CNN) with soft biometric information to improve recognition performance. Genovese \textit{et al.}~\cite{genovese2019palmnet} introduced PalmNet, an unsupervised palmprint recognition framework that integrates multiple feature extraction techniques such as CNNs, Gabor filters, and principal component analysis (PCA). Zhong \textit{et al.}~\cite{zhong2018palm} introduced a deep hashing palm vein network~(DHPN) for palm vein recognition, which can be regarded as an extension of the deep hashing network proposed in~\cite{zhu2016deep}. Recently, Yang \textit{et al.} \cite{yang2023comprehensive} proposed CCNet, which builds upon CompNet~\cite{liang2021compnet} and further enhances the competition mechanism inspired by CO3Net \cite{yang2023co}. By reformulating the traditional competition mechanism to extract diverse competitive features across channel, spatial, and multi-order dimensions, CCNet achieves strong recognition performance.

\subsection{Palmprint Recognition Methods for Open-Set Scenario}
While existing palmprint recognition methods achieve strong verification performance, variations in acquisition devices and environmental conditions induce substantial feature-distribution shifts.
To address these challenges, numerous methods have been proposed to improve robustness when recognizing previously unseen palmprint images. For example, Du \textit{et al.} \cite{du2020cross} introduced the regularized adversarial domain adaptive hashing (R-ADAH) framework for palmprint recognition. R-ADAH exhibits strong cross-domain generalization capability and does not rely on explicit identity matching between the source and target domains. 

To bridge the gap between different datasets, Shao and Zhong \cite{shao2021towards} introduced a two-stage alignment framework that performs pixel-level alignment using a style-transfer model, followed by feature-level alignment. Fei \textit{et al.} \cite{fei2023learning} proposed the cross-spectral PalmGAN, which incorporates several key constraints, including feature variance maximization and feature gap minimization, into the learning objective. This framework enables robust spectrum-invariant feature learning from multispectral palmprint images. Recently, Zhong \textit{et al.}~\cite{zhong2025regpalm} proposed RegPalm that unifies palmprint orientations and learns pairwise spatial registration in an end-to-end manner to address pattern variance in palmprint recognition.

Although current cross-device palmprint recognition methods achieve promising performance, most of them rely primarily on data augmentation strategies to improve generalization. While such approaches can alleviate domain shifts to some extent, they remain limited in explicitly handling unseen samples under open-set verification scenarios. In contrast, our framework leverages PalmBridge to map feature vectors of unseen palmprint images into a shared and structured embedding space learned from training data, thereby achieving more stable and superior verification performance on unseen samples.

\begin{figure*}[!t]
\centering
\centerline{\includegraphics[width=\textwidth]{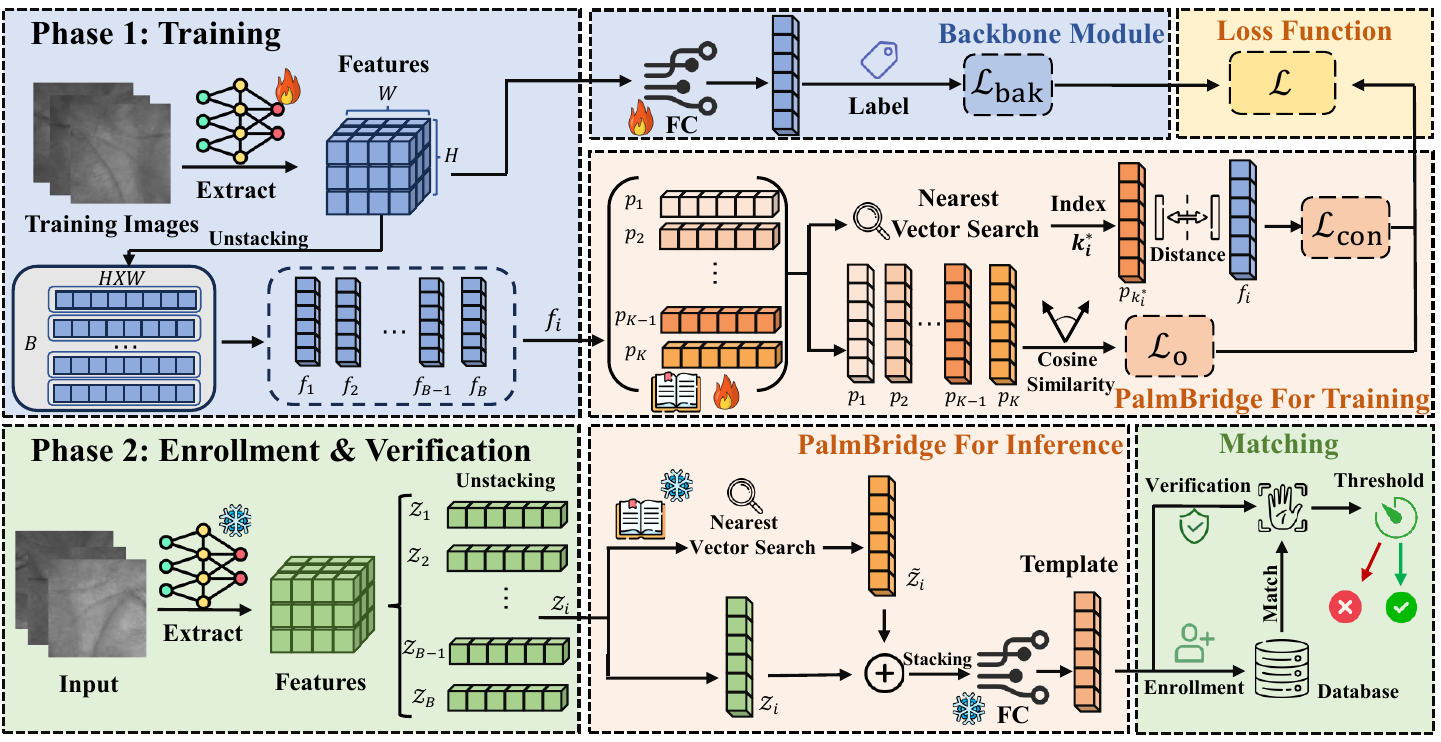}}
\caption{Pipeline of the proposed PalmBridge. During training, the PalmBridge vectors are jointly optimized with the backbone network under task-specific supervision, feature-consistency constraints, and orthogonality regularization. During enrollment and verification, PalmBridge maps extracted feature vectors into a shared embedding space to suppress nuisance variations and improve verification robustness.}
\label{fig:pipeline}
\end{figure*}

\section{Methodology}
\subsection{Overview}
As illustrated in Fig. \ref{fig:pipeline}, our framework integrates the feature-mapping module PalmBridge into the palmprint enrollment and verification pipeline. PalmBridge is inserted before the fully-connected (FC) layer and pulls the extracted feature vectors toward a shared embedding space that is already well learned by the backbone network. 
During both enrollment and verification, palmprint features are pulled toward this shared space, thereby aligning unseen query feature vectors with knowledge derived from the training data. This design effectively mitigates feature heterogeneity and substantially enhances robustness when handling previously unseen palmprint images.

\subsection{PalmBridge}
Unlike existing approaches that primarily rely on augmenting or enlarging the training data, the proposed PalmBridge focuses on reshaping query-side feature vectors, which enables the backbone network to interpret unseen samples within a shared embedding space. This design significantly reduces dependence on specialized training sets and ensures compatibility with various backbone architectures.

PalmBridge consists of a set of representative vectors denoted as $\mathbf{P} = \{p_k \mid p_k \in \mathbb{R}^D,\ k = 1,\ldots,K\}$, where
$K$ denotes the number of representative vectors, and $D$ represents the feature dimension. $p_k$ represents the $k$-th vector in the PalmBridge set $\mathbf{P}$.

The core of PalmBridge lies in a feature-mapping procedure that aligns the feature vectors extracted by the backbone with the learned representative vectors. Given a batch of palmprint feature vectors $\mathbf{z}\in\mathbb{R}^{B\times D}$ extracted by the backbone network, PalmBridge computes the distances between each feature vector $z_i$ and all representative vectors, assigning each feature vector to its nearest vector. The process is defined as follows:
\begin{equation}
    k_i^*=\arg\min_k\|z_i-p_k\|_2^2 \quad k\in\left[1,K\right],
\end{equation}
where $k_i^*$ is the index of the representative vector closest to $z_i$. The mapped representation is defined by:
\begin{equation}
\tilde{z}_i=p_{k_i^*} \quad i\in\left[1,B\right],
\end{equation}
where $z_i$ replaces itself with $\tilde{z}_i$ according to the index of PalmBridge vectors, thus forming a batch of mapped feature vectors $\tilde{\mathbf{z}}$, which is described as follows:
\begin{equation}
    \tilde{\mathbf{z}}=[\tilde{z}_{1};\ldots;\tilde{z}_{B}]\quad \tilde{z}_{i}\in\mathbb{R}^D,
\end{equation}
where $i$ is an index variable ranging from $1$ to $B$, such that $\tilde{z}_{i}$ represents corresponding mapped vector of $z_i$ in $\textbf{z}$.

To preserve identity-specific information during mapping, PalmBridge blends the original feature vector with its mapped counterpart via a linear combination:
\begin{equation}
    \hat{z}_i= w_{ori} \times z_i + w_{map} \times \tilde{z}_i \quad i\in\left[1,B\right],
    \label{eq:weighted mapped}
\end{equation}
where $w_{ori}$ and $w_{map}$ are blending coefficients controlling the contributions of the original vector and mapped vector. respectively. thereby yielding a batch of the final feature vectors $\hat{\mathbf{z}}$:
\begin{equation}
    \hat{\mathbf{z}}=[\hat{z}_{1};\ldots;\hat{z}_{B}]\quad \hat{z}_{i} \in\mathbb{R}^{D},
\end{equation}
where $i$ is an index variable ranging from $1$ to $B$, such that $\hat{z}_{i}$ represents corresponding final feature vector of $z_i$ in $\textbf{z}$.

Through this linear blending operation, feature vectors are projected toward a shared embedding space while retaining identity-discriminative information from the original vectors. As a result, the final feature vector is aligned within the shared embedding space while preserving the distinctive characteristics of each individual vector.

\subsection{Loss functions}
\subsubsection{Feature Consistency Loss}
We design a feature consistency loss $\mathcal{L}_{\mathrm{con}}$ to stabilize the joint optimization of the PalmBridge vectors and the backbone feature extractor, while ensuring that the learned vectors remain discriminative.
The loss is implemented as a symmetric consistency constraint that enforces mutual alignment between the mapped vectors in PalmBridge and the features extracted by the backbone. Specifically, $\mathcal{L}_{\mathrm{con}}$ is formulated as follows:
\begin{equation}
\mathcal{L}_{\mathrm{con}} =
\frac{1}{B} \sum_{i=1}^{B}
\Big(
\left\| p_i - \mathrm{sg}(z_i) \right\|_2^2
+
\lambda\left\| \mathrm{sg}(p_i) - z_i \right\|_2^2
\Big),
\end{equation}
where $z_i$ denotes the feature vector extracted by the backbone network, and $p_i$ denotes the corresponding mapped vector in PalmBridge. The operator $\mathrm{sg}(\cdot)$ stops gradient propagation during back-propagation. The hyperparameter $\lambda$ controls the strength of the second term and is empirically set to 0.25 in this paper.

The first term aligns PalmBridge vectors with backbone features by updating the mapped vectors while preventing gradients from flowing into the backbone. Conversely, the second term enforces backbone features to stay close to their assigned mapped vectors, while blocking gradients from updating the PalmBridge vectors. Together, these two terms enforce a symmetric consistency constraint between PalmBridge vectors and backbone features.

\subsubsection{Orthogonal Loss}
To enhance the vector diversity in PalmBridge and reduce representational redundancy, we incorporate the orthogonal loss $\mathcal{L}_{\mathrm{o}}$. This loss encourages PalmBridge vectors to remain mutually decorrelated by penalizing non-orthogonal relationships, which promote a more diverse and well-structured feature space. 

Let $\textbf{W}\in\mathbb{R}^{K\times D}$ denote the matrix of normalized PalmBridge vectors, where each row corresponds to an $\ell_2$-normalized vector. The pairwise similarity matrix $\textbf{S}\in \mathbb{R}^{K\times K}$ is defined as $\mathbf{S} = \mathbf{W}\mathbf{W}^{\top}$.
Then, the orthogonality loss is formulated as follows:
\begin{equation}
\mathcal{L}_{\mathrm{o}}
= \frac{1}{K^{2}} 
\sum_{i=1}^{K} \sum_{j=1}^{K}
\left( S_{ij} - \delta_{ij} \right)^{2},
\label{eq:orthogonal}
\end{equation}
where $\delta_{ij}$ is the Kronecker delta, $S_{ij}$ in $\mathbf{S}$ denotes the cosine similarity between normalized ${p}_{i}$ and ${p}_{j}$ in $\textbf{P}$. 

By encouraging the similarity matrix $\mathbf{S}$ to approach the identity matrix, $\mathbf{S}$ suppresses correlations among different feature vectors and reduces feature redundancy.



\subsubsection{Total Loss}
In addition to the losses mentioned above, a task-specific loss $\mathcal{L}_\text{bak}$ is required to optimize the backbone network. Notably, PalmBridge is designed as a plug-and-play module, and thus $\mathcal{L}_\text{bak}$ can take different forms depending on the specific backbone architecture or learning paradigm employed. The overall training objective is formulated as follows:
\begin{equation}
    \mathcal{L} = \mathcal{L}_{\mathrm{bak}} + \alpha\mathcal{L}_{\mathrm{con}} + \beta\mathcal{L}_{\mathrm{o}},
    \label{eq:total loss}
\end{equation}
where $\alpha$ and $\beta$ are the weights of feature consistency loss and orthogonal loss, respectively, both empirically set to 1 in this paper.


\subsection{Theoritical Analysis}
To facilitate a clearer understanding of the proposed framework, we present a theoretical analysis in this section. The conventional feature extraction process of a palmprint backbone can be modeled as:
\begin{equation}
    z = f_\theta(x)=\mu_{y}+\epsilon(x),
\end{equation}
where $\mu_{y}$ is an identity-specific latent vector and $\epsilon(x)$ captures nuisance (verification-
irrelevant) variation and acquisition-induced heterogeneity.
Under an open-set scenario, the backbone is optimized using identities $y\in\mathcal{J}_{\mathrm{train}}$. When encountering unseen identities $y\in\mathcal{J}_{\mathrm{test}}$, the nuisance term $\epsilon(x)$ typically increases, which leads to inflated within-identity dispersion in the feature space. 

Then, the PalmBridge mapping can be formulated as:
\begin{equation}
    T_{\alpha}(z) = (1-\alpha)z + \alpha\Pi_{p}(z),
\end{equation}
\begin{equation}
    \Pi_{p}(z) = p_{\arg\min_k\|z-p_k\|_2^2},
\end{equation}
where $\alpha=w_{map}$ and $1-\alpha=w_{ori}$. If the learned vector set $\mathbf{P}$ approximates the high-density regions of the training feature distribution, the operator $\Pi_{p}$ effectively acts as a learned denoiser, pulling test feature vectors toward a stable set of learned vectors.

Importantly, the performance gains of PalmBridge in open-set verification do not arise from introducing new identity information, but rather from suppressing nuisance variance more strongly than degrading identity separability. This mechanism leads to consistent improvements in open-set performance.
To illustrate this effect, consider two samples from the same unseen identity: $z_1 = \mu + \epsilon_1$ and $z_2 = \mu + \epsilon_2$. If both features are assigned to the same Voronoi region induced by $\mathbf{P}$, i.e., $\Pi_p(z_1)=\Pi_p(z_2)$, the squared distance between the mapped features becomes:
\begin{equation}
    \|T_{\alpha}(z_1)-T_{\alpha}(z_2)\|_2^2  = (1-\alpha)\|\epsilon_1-\epsilon_2\|_2^2,
\end{equation}

Thus, the genuine-pair dispersion is reduced by a factor of $1-\alpha$ whenever two same-identity features share the same representative region. In practice, the effectiveness of this contraction depends on the assignment consistency
\begin{equation}
p_{\text{same}}=\Pr\!\big(\Pi_{p}(z_{1})=\Pi_{p}(z_{2}) \mid y_{1}=y_{2}\big),
\end{equation}
which measures how often genuine pairs are mapped to the same learned vector. In contrast, for two features from different identities $\mu_a \neq \mu_b$, the mapped difference is given by
\begin{equation}
T_{\alpha}(z_{a})-T_{\alpha}(z_{b})=\alpha(p_{a}-p_{b})+(1-\alpha)(z_{a}-z_{b}).
\end{equation}
This separation behavior in this case is governed by the assignment collision rate:
\begin{equation}
p_{\text{collide}}=\Pr\!\big(\Pi_{p}(z_{a})=\Pi_{p}(z_{b}) \mid y_{a}\neq y_{b}\big),
\end{equation}
since large $\alpha$ can increase the risk of identity mixing when $p_{\text{collide}}$ is non-negligible. Accordingly, PalmBridge effectively reduces genuine-pair dispersion when $p_{\text{same}}$ is high, while preserving inter-class separation when $p_{\text{collide}}$ remains low. This analysis clarifies the observed performance trade-off controlled by the blending coefficient $\alpha$, and provides theoretical insight into the robustness gains achieved by PalmBridge in open-set palmprint verification.

\begin{table*}[!t]
\centering
\caption{Intra-dataset open-set results on the IITD dataset.}
\resizebox{\textwidth}{!}{%
\begin{tabular}{lcccccccccccc} 
\hline
\multirow{2}{*}{Framework} 
& \multicolumn{2}{c}{ResNet18\cite{he2016deep}} & \multicolumn{2}{c}{DHN\cite{wu2021palmprint}} & \multicolumn{2}{c}{CompNet\cite{liang2021compnet}}
& \multicolumn{2}{c}{CCNet\cite{yang2023comprehensive}} & \multicolumn{2}{c}{CO3Net\cite{yang2023co}} & \multicolumn{2}{c}{SF2Net\cite{liu2025sf2net}} \\
\cmidrule(lr){2-3}\cmidrule(lr){4-5}\cmidrule(lr){6-7}
\cmidrule(lr){8-9}\cmidrule(lr){10-11}\cmidrule(lr){12-13}
& ACC(\%)$\uparrow$ & EER(\%)$\downarrow$ 
& ACC(\%)$\uparrow$ & EER(\%)$\downarrow$ 
& ACC(\%)$\uparrow$ & EER(\%)$\downarrow$
& ACC(\%)$\uparrow$ & EER(\%)$\downarrow$
& ACC(\%)$\uparrow$ & EER(\%)$\downarrow$
& ACC(\%)$\uparrow$ & EER(\%)$\downarrow$ \\ \hline

Naive 
& 84.78 & 7.3188   %
& 86.96 & 6.5217  %
& 98.26 & 2.0290   %
& 97.39 & 2.2679  %
& 97.82 & 2.6812   %
& 98.26 & 2.3188   %
\\

PalmRSS
& 86.52 & 6.0145   %
& 80.87 & 8.6957   %
& 99.13 & 2.0290   %
& 97.61 & 2.9710   %
& 98.26 & 2.7536   %
& 98.04 & 1.8841   %
\\

C-LMCL
& 91.96 & 4.6377   %
& 90.44 & 6.5876   %
& 98.48 & 2.1739   %
& \textbf{98.48} & 2.1739   %
& 98.04 & 2.6812   %
& 98.04 & 2.8936   %
\\

UAA
& 86.74 & 8.0272   %
& 57.39 & 21.6667   %
& 97.61 & 4.2029   %
& 98.26 & 2.4435   %
& 96.74 & 3.9130   %
& 98.70 & 2.3188   %
\\

\cellcolor{lightergray}\textbf{PalmBridge} 
& \cellcolor{lightergray}\textbf{92.83} & \cellcolor{lightergray}\textbf{4.4228}
& \cellcolor{lightergray}\textbf{90.44} & \cellcolor{lightergray}\textbf{5.1449}
& \cellcolor{lightergray}\textbf{99.13} & \cellcolor{lightergray}\textbf{1.3534}
& \cellcolor{lightergray}98.04& \cellcolor{lightergray}\textbf{2.0689}
& \cellcolor{lightergray}\textbf{98.26}& \cellcolor{lightergray}\textbf{2.1739}
& \cellcolor{lightergray}\textbf{98.70} & \cellcolor{lightergray}\textbf{1.7391}
\\

\hline
\end{tabular}
}
\label{tab:IntraIITD}
\end{table*}

\begin{table*}[!t]
\centering
\caption{Intra-dataset open-set results on the PolyU dataset.}
\resizebox{\textwidth}{!}{%
\begin{tabular}{lcccccccccccc} 
\hline
\multirow{2}{*}{Framework} 
& \multicolumn{2}{c}{ResNet18\cite{he2016deep}} & \multicolumn{2}{c}{DHN\cite{wu2021palmprint}} & \multicolumn{2}{c}{CompNet\cite{liang2021compnet}}
& \multicolumn{2}{c}{CCNet\cite{yang2023comprehensive}} & \multicolumn{2}{c}{CO3Net\cite{yang2023co}} & \multicolumn{2}{c}{SF2Net\cite{liu2025sf2net}} \\
\cmidrule(lr){2-3}\cmidrule(lr){4-5}\cmidrule(lr){6-7}
\cmidrule(lr){8-9}\cmidrule(lr){10-11}\cmidrule(lr){12-13}
& ACC(\%)$\uparrow$ & EER(\%)$\downarrow$ 
& ACC(\%)$\uparrow$ & EER(\%)$\downarrow$ 
& ACC(\%)$\uparrow$ & EER(\%)$\downarrow$
& ACC(\%)$\uparrow$ & EER(\%)$\downarrow$
& ACC(\%)$\uparrow$ & EER(\%)$\downarrow$
& ACC(\%)$\uparrow$ & EER(\%)$\downarrow$ \\ \hline

Naive 
& 68.25 & 9.5714   %
& 66.08 & 9.0488  %
& 98.84 & 1.0741   %
& 98.57 & 1.7831  %
& 98.10 & 1.8385   %
& 98.89 & 1.4921   %
\\

PalmRSS
& 49.00 & 13.8317   %
& 47.67 & 15.4021   %
& 99.05 & 1.1574   %
& 98.04 & 2.1029   %
& \textbf{99.31} & 1.4678   %
& 99.21 & \textbf{0.9939}   %
\\

C-LMCL
& 61.22 & 12.4998   %
& 61.69 & 10.7562   %
& 97.99 & 1.7407   %
& 98.68 & 1.4868   %
& 98.31 & 1.6190   %
& 99.26 & 1.3436   %
\\

UAA
& 49.05 & 18.4499   %
& 47.20 & 17.4709   %
& 99.21 & 1.4815   %
& 98.73 & 1.4709   %
& 98.15 & 1.7989   %
& \textbf{99.31} & 1.1217   %
\\

\cellcolor{lightergray}\textbf{PalmBridge} 
& \cellcolor{lightergray}\textbf{69.10} & \cellcolor{lightergray}\textbf{9.4646}
& \cellcolor{lightergray}\textbf{67.83} & \cellcolor{lightergray}\textbf{8.9101}
& \cellcolor{lightergray}\textbf{99.21} & \cellcolor{lightergray}\textbf{0.8254}
& \cellcolor{lightergray}\textbf{98.78}& \cellcolor{lightergray}\textbf{1.2222}
& \cellcolor{lightergray}98.52& \cellcolor{lightergray}\textbf{1.4220}
& \cellcolor{lightergray}98.52 & \cellcolor{lightergray}1.4550
\\

\hline
\end{tabular}
}
\label{tab:IntraPolyU}
\end{table*}

\begin{table*}[!t]
\centering
\caption{Intra-dataset open-set results on the Tongji dataset.}
\resizebox{\textwidth}{!}{%
\begin{tabular}{lcccccccccccc} 
\hline
\multirow{2}{*}{Framework} 
& \multicolumn{2}{c}{ResNet18\cite{he2016deep}} & \multicolumn{2}{c}{DHN\cite{wu2021palmprint}} & \multicolumn{2}{c}{CompNet\cite{liang2021compnet}}
& \multicolumn{2}{c}{CCNet\cite{yang2023comprehensive}} & \multicolumn{2}{c}{CO3Net\cite{yang2023co}} & \multicolumn{2}{c}{SF2Net\cite{liu2025sf2net}} \\
\cmidrule(lr){2-3}\cmidrule(lr){4-5}\cmidrule(lr){6-7}
\cmidrule(lr){8-9}\cmidrule(lr){10-11}\cmidrule(lr){12-13}
& ACC(\%)$\uparrow$ & EER(\%)$\downarrow$ 
& ACC(\%)$\uparrow$ & EER(\%)$\downarrow$ 
& ACC(\%)$\uparrow$ & EER(\%)$\downarrow$
& ACC(\%)$\uparrow$ & EER(\%)$\downarrow$
& ACC(\%)$\uparrow$ & EER(\%)$\downarrow$
& ACC(\%)$\uparrow$ & EER(\%)$\downarrow$ \\ \hline

Naive 
& 96.00 & 5.2638   %
& 94.60 & 5.9554  %
& 100.00 & 0.1733   %
& 100.00 & 0.2131  %
& 100.00 & 0.2000   %
& 99.87 & 0.1867   %
\\

PalmRSS
& 95.13 & 5.2009   %
& 94.47 & 5.7600   %
& 100.00 & 0.1467   %
& 99.93 & 0.1885   %
& 100.00 & 0.2667   %
& 100.00 & 0.1867   %
\\

C-LMCL
& 97.00 & 4.5867   %
& 96.00 & 5.0667   %
& 100.00 & 0.2126   %
& 99.93 & 0.2800   %
& 99.93 & 0.2000   %
& 99.93 & 0.2800   %
\\

UAA
& 90.27 & 12.6267   %
& 92.73 & 6.5718   %
& 100.00 & 0.2238   %
& 100.00 & 0.2267   %
& 100.00 & 0.3687   %
& 100.00 & 0.2667   %
\\

\cellcolor{lightergray}\textbf{PalmBridge} 
& \cellcolor{lightergray}\textbf{97.40} & \cellcolor{lightergray}\textbf{3.6800}
& \cellcolor{lightergray}\textbf{97.80} & \cellcolor{lightergray}\textbf{3.3844}
& \cellcolor{lightergray}\textbf{100.00} & \cellcolor{lightergray}\textbf{0.1333}
& \cellcolor{lightergray}\textbf{100.00}& \cellcolor{lightergray}\textbf{0.1333}
& \cellcolor{lightergray}\textbf{100.00}& \cellcolor{lightergray}\textbf{0.1733}
& \cellcolor{lightergray}\textbf{100.00} & \cellcolor{lightergray}\textbf{0.1733}
\\

\hline
\end{tabular}
}
\label{tab:IntraTongji}
\end{table*}

\begin{table*}[!t]
\centering
\caption{Intra-dataset open-set results on the palmvein dataset.}
\resizebox{\textwidth}{!}{%
\begin{tabular}{lcccccccccccc} 
\hline
\multirow{2}{*}{Framework} 
& \multicolumn{2}{c}{ResNet18\cite{he2016deep}} & \multicolumn{2}{c}{DHN\cite{wu2021palmprint}} & \multicolumn{2}{c}{CompNet\cite{liang2021compnet}}
& \multicolumn{2}{c}{CCNet\cite{yang2023comprehensive}} & \multicolumn{2}{c}{CO3Net\cite{yang2023co}} & \multicolumn{2}{c}{SF2Net\cite{liu2025sf2net}} \\
\cmidrule(lr){2-3}\cmidrule(lr){4-5}\cmidrule(lr){6-7}
\cmidrule(lr){8-9}\cmidrule(lr){10-11}\cmidrule(lr){12-13}
& ACC(\%)$\uparrow$ & EER(\%)$\downarrow$ 
& ACC(\%)$\uparrow$ & EER(\%)$\downarrow$ 
& ACC(\%)$\uparrow$ & EER(\%)$\downarrow$
& ACC(\%)$\uparrow$ & EER(\%)$\downarrow$
& ACC(\%)$\uparrow$ & EER(\%)$\downarrow$
& ACC(\%)$\uparrow$ & EER(\%)$\downarrow$ \\ \hline

Naive 
& 74.87 & 9.3778   %
& 71.47 & 11.9667  %
& 99.87 & 0.4111   %
& 99.93 & 0.5577  %
& 99.93 & 0.4481   %
& 99.87 & 0.3444   %
\\

PalmRSS
& 74.80 & 8.3129   %
& 70.13 & 9.5200   %
& 99.87 & 0.4556   %
& 99.87 & 0.5735   %
& 99.93 & 0.4055   %
& 99.93 & 0.3556   %
\\

C-LMCL
& \textbf{85.67} & \textbf{6.6333}   %
& \textbf{80.87} & \textbf{8.0000}   %
& 99.80 & 0.3356   %
& 99.80 & 0.5289   %
& 99.93 & 0.4058   %
& 99.93 & 0.3300   %
\\

UAA
& 70.60 & 15.7860   %
& 71.80 & 10.2759   %
& 99.73 & 0.4556   %
& 99.93 & 0.3760   %
& 99.93 & 0.5333   %
& 99.93 & \textbf{0.2667}   %
\\

\cellcolor{lightergray}\textbf{PalmBridge} 
& \cellcolor{lightergray}83.47 & \cellcolor{lightergray}7.0778
& \cellcolor{lightergray}78.80 & \cellcolor{lightergray}8.7651
& \cellcolor{lightergray}\textbf{99.87} & \cellcolor{lightergray}\textbf{0.2778}
& \cellcolor{lightergray}\textbf{100.00}& \cellcolor{lightergray}\textbf{0.3010}
& \cellcolor{lightergray}\textbf{100.00}& \cellcolor{lightergray}\textbf{0.3667}
& \cellcolor{lightergray}\textbf{99.93} & \cellcolor{lightergray}0.2889
\\

\hline
\end{tabular}
}
\label{tab:IntraNIR}
\end{table*}

\begin{table*}[!t]
\centering
\caption{Results of the cross-dataset open-set experiments.}
\begin{tabular}{llcccccc}
\hline
\multirow{2}{*}{Metrics}               & \multirow{2}{*}{Framework}   & \multicolumn{2}{c}{IITD} & \multicolumn{2}{c}{PolyU} & \multicolumn{2}{c}{Tongji} \\ \cmidrule(lr){3-4} \cmidrule(lr){5-6} \cmidrule(lr){7-8} 
             &     & PolyU        & Tongji       & IITD           & Tongji         & IITD           & PolyU            \\ \hline

\multirow{3}{*}{EER(\%)$\downarrow$} & Naive &            2.7646&           3.1000&             6.3043&            2.2561&             6.2319&            1.6534          \\
&\cellcolor{lightergray}\textbf{PalmBridge} &\cellcolor{lightergray}2.3435&\cellcolor{lightergray}2.8791&\cellcolor{lightergray}6.2059&\cellcolor{lightergray}2.2457&\cellcolor{lightergray}6.0870     &\cellcolor{lightergray}1.6111  \\
&\cellcolor{lightergray}$\triangle$ &\cellcolor{lightergray}\textcolor{red}{-0.4211}&\cellcolor{lightergray}\textcolor{red}{-0.2209}&\cellcolor{lightergray}\textcolor{red}{-0.0984}&\cellcolor{lightergray}\textcolor{red}{-0.0104} &\cellcolor{lightergray}\textcolor{red}{-0.1449}       &\cellcolor{lightergray}\textcolor{red}{-0.0423} 
\\ \hline
\multirow{3}{*}{ACC(\%)$\uparrow$} & Naive &             99.55&           98.53&              95.98&            98.17&             96.74&           99.76           \\
&\cellcolor{lightergray}\textbf{PalmBridge} &\cellcolor{lightergray}99.63&\cellcolor{lightergray}98.35&\cellcolor{lightergray}96.30&\cellcolor{lightergray}98.55&\cellcolor{lightergray}96.74       &\cellcolor{lightergray}99.66  \\
&\cellcolor{lightergray}$\triangle$ &\cellcolor{lightergray}\textcolor{red}{+0.08}&\cellcolor{lightergray}\textcolor{blue}{-0.18}&\cellcolor{lightergray}\textcolor{red}{+0.32}&\cellcolor{lightergray}\textcolor{red}{+0.38} &\cellcolor{lightergray}\textcolor{red}{+0.00}       &\cellcolor{lightergray}\textcolor{blue}{-0.10} 
\\ \hline
\end{tabular}
\label{tab:Cross}
\end{table*}

\renewcommand{\arraystretch}{1.4}   %
\begin{table}[!t]
\centering
\caption{EERs of the closed-set experiments on public datasets.}
\begin{tabular}{lccccc}
\hline
Method & PolyU &Tongji&IITD \\ \hline

PalmCode\cite{zhang2003online}   & 0.35000 & 0.11000 &5.45 
\\ \hline
Fusion Code\cite{kong2006palmprint}   & 0.24000 & 0.07310 &6.20 
\\ \hline
Comp Code\cite{kong2004competitive}   & 0.12000 & 0.11000 &5.50 
\\ \hline
RLOC\cite{jia2008palmprint}   & 0.13000   & 0.02530 &5.00 
\\ \hline
BOCV\cite{guo2009palmprint}     & 0.08130  & 0.00560 &4.56 
\\ \hline
E-BOCV\cite{zhang2012fragile}    & 0.09950  & 0.01800 &4.65 
\\ \hline
HOC\cite{fei2016half}    & 0.16000  & 0.09540 &6.55 
\\ \hline
DCC\cite{xu2018discriminative}     & 0.15000   & 0.05060 &5.49 
\\ \hline
DRCC\cite{xu2018discriminative}     & 0.18000   & 0.03080 &5.42 
\\ \hline
DHPN\cite{zhong2018palm}    & 0.03200  & 0.06590 &3.73 
\\ \hline
PalmNet\cite{genovese2019palmnet}     & 0.11100  & 0.03220 &4.20 
\\ \hline
EDM\cite{yang2020extreme}     & 0.06090   & 0.01130 &4.56 
\\ \hline
DHN\cite{wu2021palmprint}    & 0.03720 & 0.08790 &4.30
\\ \hline
2TCC\cite{yang2023multi}     & 0.08340 & 0.00750 &5.94 
\\ \hline
MTCC\cite{yang2023multi}     & 0.05490  & 0.00430 &3.94 
\\ \hline
CompNet\cite{liang2021compnet}    & 0.05560   & 0.02500 &0.54 
\\ \hline
CO3Net\cite{yang2023co}           & 0.02200 & 0.00500 &0.47 \\ \hline
CCNet\cite{yang2023comprehensive}  & 0.00006 & 0.00004 &0.18 \\ \hline

\cellcolor{lightergray}\textbf{PalmBridge} 
& \cellcolor{lightergray}\textbf{0}
& \cellcolor{lightergray}\textbf{0.00833}
& \cellcolor{lightergray}\textbf{0.36}
\\ \hline
\end{tabular}
\label{tab:Close}
\end{table}

\begin{table}[!t]
\centering
\caption{EERs of the closed experiments on the palmvein dataset.}
\begin{tabular}{lc} %
\hline
Method & EER(\%)$\downarrow$\\ \hline

PalmCode\cite{zhang2003online} & 0.20000 
 \\ \hline
Fusion Code\cite{kong2006palmprint} & 0.17000 
 \\ \hline
Comp Code\cite{kong2004competitive} & 0.05790 
 \\ \hline
RLOC\cite{jia2008palmprint} & 0.06290 
 \\ \hline
BOCV\cite{guo2009palmprint} & 0.02610 
 \\ \hline
E-BOCV\cite{zhang2012fragile} & 0.03150 
 \\ \hline
HOC\cite{fei2016half} & 0.08390 
 \\ \hline
DCC\cite{xu2018discriminative} & 0.05750 
 \\ \hline
DRCC\cite{xu2018discriminative} & 0.05630 
 \\ \hline
DHPN\cite{zhong2018palm} & 0.00200 
 \\ \hline
PalmNet\cite{genovese2019palmnet} & 0.08710 
 \\ \hline
EDM\cite{yang2020extreme} & 0.03630 
 \\ \hline
DHN\cite{wu2021palmprint} & 0.02330 
 \\ \hline
2TCC\cite{yang2023multi} & 0.03220 
 \\ \hline
MTCC\cite{yang2023multi} & 0.00950 
 \\ \hline
CompNet\cite{liang2021compnet} & 0.00250 
 \\ \hline
CO3Net\cite{yang2023co} & 0.00050 
 \\ \hline
CCNet\cite{yang2023comprehensive} & 0 
 \\ \hline

\cellcolor{lightergray}\textbf{PalmBridge} 
& \cellcolor{lightergray}\textbf{0.00069}
 \\ \hline

\end{tabular}
\label{tab:CloseNIR}
\end{table}

\section{Experiments and Discussions}
\subsection{Experimental Settings}
\subsubsection{Datasets}
We validated the proposed PalmBridge framework on four palmprint datasets, including both contact-based and contactless palmprint datasets, as well as a palmvein dataset.
The datasets used are summarized as follows:

PolyU \cite{zhang2003online} was constructed using a contact-based palmprint acquisition device and includes samples from 193 subjects. Data collection occurred in two sessions, approximately two months apart.
The complete dataset comprises 7,752 images, with 7,560 from 189 subjects (378 palms), typically used for experimental evaluation.

Tongji \cite{zhang2017towards} was captured from 300 participants using a contactless acquisition device.
Data collection was conducted in two sessions, separated by about two months. Each participant contributed 20 palm images per session, yielding a total of 12,000 samples from 600 distinct palms.

IITD \cite{kumar2010personal} was acquired from 460 subjects using a contactless palmprint imaging system. Approximately five samples were captured from each individual, yielding a total of 2,601 images. After preprocessing and quality assessment, 2,300 images were retained for experimental evaluation.

PalmVein~\cite{zhang2009online} was collected using a near-infrared (NIR) imaging system as part of the Multi-Spectral palmprint dataset. The NIR subset captures palm vein patterns under near-infrared illumination and is used in this work for experimental evaluation of palm vein recognition.

\subsubsection{Backbones}
We first adopt the classic ResNet-18 \cite{he2016deep} as a baseline backbone to evaluate the effectiveness of the proposed method. Moreover, we choose five representative palmprint recognition models, including DHN~\cite{wu2021palmprint}, CompNet~\cite{liang2021compnet}, CO3Net~\cite{yang2023co}, CCNet~\cite{yang2023comprehensive}, and SF2Net~\cite{liu2025sf2net}, to comprehensively validate the generality and effectiveness of PalmBridge across diverse architectures. 

\subsubsection{Baselines}
We compared the proposed method with several training paradigms, including a naive baseline without data augmentation, feature alignment, or regularization, as well as PalmRSS~\cite{jia2025single}, C-LMCL~\cite{zhong2019centralized}, and UAA~\cite{jin2025unified}.

\subsubsection{Implementation Details}
PalmBridge is implemented with the PyTorch framework. During training, we employ the Adam optimizer with a learning rate of 0.001. The batch size is set to 16 for all models, except for SF2Net, which uses a batch size of 500. All experiments are conducted on a single NVIDIA RTX 3090 GPU.

\subsubsection{Evaluating Metrics}
We adopt equal error rate (EER) and rank-1 accuracy (ACC) as the evaluation metrics. EER, a widely used measure in biometric verification, is defined at the operating point where the false acceptance rate (FAR) equals the false rejection rate (FRR), with lower values indicating better performance. In addition, receiver operating characteristic~ (ROC) curves and the distributions of genuine and impostor (GI) similarity scores are analyzed to further assess feature separability and model reliability.

\subsubsection{Experimental Implementation}
We design three types of experimental protocols to comprehensively evaluate the effectiveness of PalmBridge, including intra-dataset open-set,cross-dataset open-set, and closed-set verification scenarios.

\textbf{Intra-dataset open-set evaluation.} For each dataset, all identities are evenly divided into two groups with no identity overlap. In the first group, images of each identity are further split into training and testing subsets for model optimization and internal evaluation. In the second group, images are equally partitioned into gallery and query subsets for open-set verification. This protocol enables open-set evaluation under a controlled and consistent experimental setting.

\textbf{Cross-dataset open-set evaluation.} To assess generalization across heterogeneous acquisition conditions, the IITD, PolyU, and Tongji datasets are employed in a rotational training–testing scheme. In each evaluation round, one dataset is used exclusively for model training and validation, while the remaining two datasets are reserved solely for gallery–query verification. This setup enables a rigorous assessment of cross-dataset generalization performance.

\textbf{Closed-set evaluation.} In the closed-set scenario, all identities within each dataset are preserved without cross-dataset separation. For instance, in the PolyU dataset, all 378 identities are included, with the first 10 images per identity used for training, and the remaining images used for testing and verification.

\begin{figure}[!t]
    \centering
    \begin{minipage}[t]{0.48\linewidth}
        \centering
        \includegraphics[width=\linewidth]{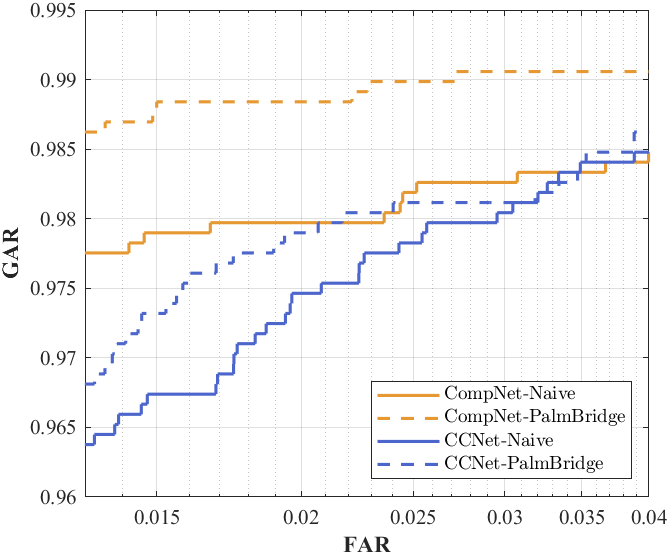}
        \centerline{(a)}
    \end{minipage}
    \hfill
    \begin{minipage}[t]{0.48\linewidth}
        \centering
        \includegraphics[width=\linewidth]{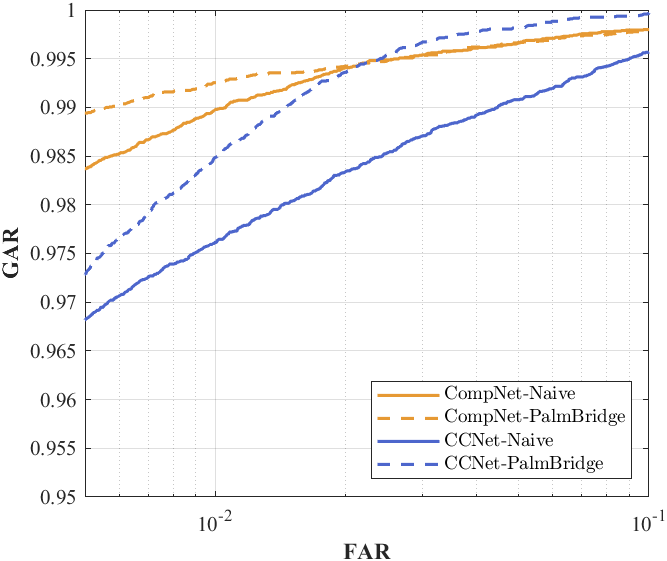}
        \centerline{(b)}
    \end{minipage}


    \begin{minipage}[t]{0.48\linewidth}
        \centering
        \includegraphics[width=\linewidth]{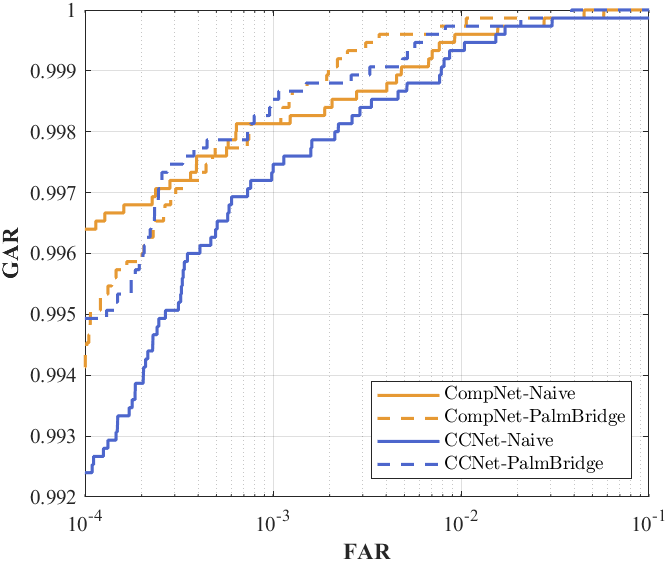}
        \centerline{(c)}
    \end{minipage}
    \hfill
    \begin{minipage}[t]{0.48\linewidth}
        \centering
        \includegraphics[width=\linewidth]{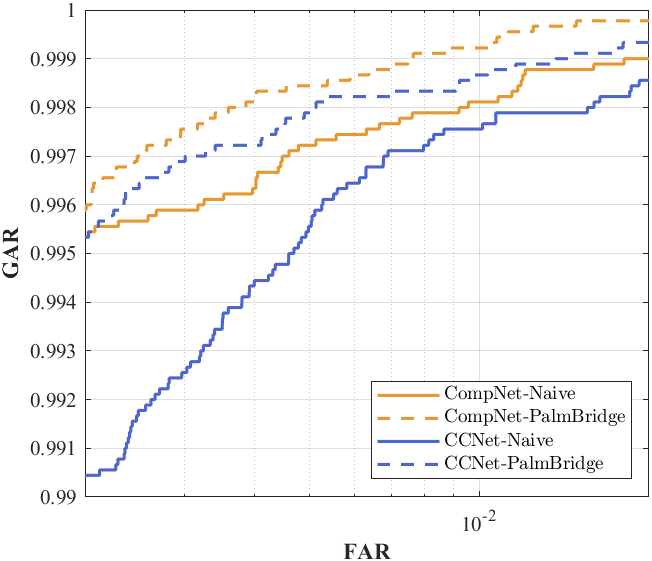}
        \centerline{(d)}
    \end{minipage}

    \caption{ROC curves of CompNet and CCNet across different datasets. (a)-(d) represent the ROC curves on IITD, PolyU, Tongji, and PalmVein, respectively, comparing the naive baseline and the PalmBridge-enhanced frameworks for both CompNet and CCNet.}
    \label{fig:roc}
\end{figure}

\begin{figure}[!t]
	\centering
	\begin{minipage}[t]{0.49\columnwidth}
	\centering
	\includegraphics[width=\textwidth]{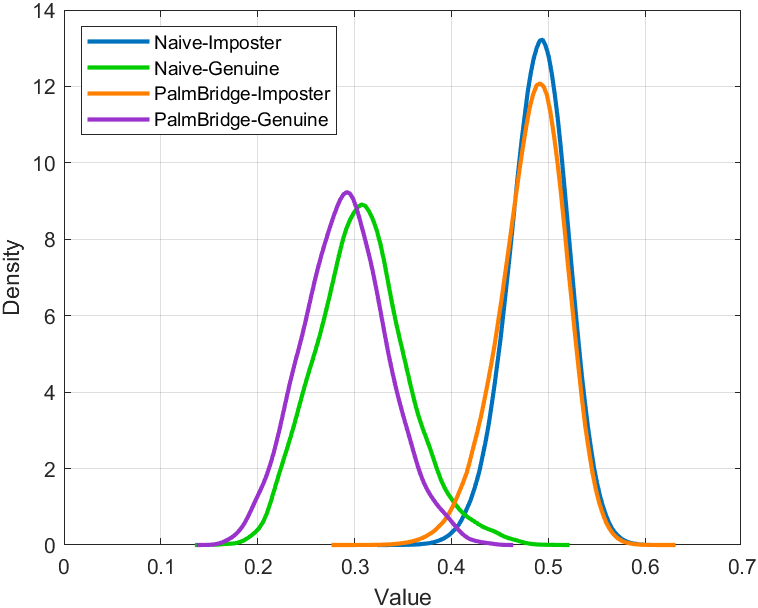}
	\centerline{(a)}
	\end{minipage}
	\begin{minipage}[t]{0.49\columnwidth}
	\includegraphics[width=\textwidth]{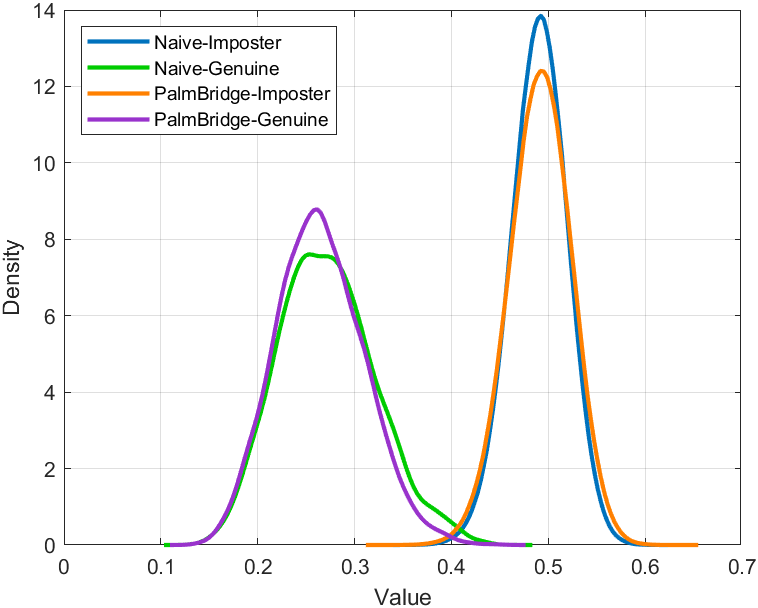}
	\centerline{(b)}
	\end{minipage}
\caption{Genuine-Imposter curves of CCNet on different datasets. (a) and (b) represent the GI curves on PolyU and PalmVein, respectively, comparing the naive framework and the proposed PalmBridge-enhanced framework.}
\label{fig:GI}
\end{figure}

\subsection{Intra-Dataset Open-Set Experiments}
We first evaluate the performance of the proposed PalmBridge under the intra-dataset open-set scenario, and the results are reported in Tabs. \ref{tab:IntraIITD}-\ref{tab:IntraNIR}. 
Overall, PalmBridge consistently outperforms the naive baseline as well as PalmRSS, C-LMCL, and UAA across most backbone networks and datasets. In particular, PalmBridge achieves higher ACC and lower EER, which demonstrates its superior generalization capability under intra-dataset open-set scenarios. While other frameworks provide moderate improvements over the naive baseline on certain backbones, PalmBridge yields more consistent and substantially larger gains, especially in EER reduction.

Notable performance improvements are observed for CompNet and CCNet, where the integration of PalmBridge leads to consistent gains across all evaluated palmprint datasets. For example, on IITD, PalmBridge improves CCNet by 0.65\% in ACC and reduces the EER by 0.5609\% on PolyU, corresponding to nearly a 30\% improvement over the naive framework and outperforming all competing frameworks. Similarly, when applied to CompNet with IITD, PalmBridge achieves an ACC increase of 0.87\% and an EER reduction of 0.6756\%, resulting in approximately a 35\% gain over the naive baseline and a clear advantage over other frameworks. These results suggest that PalmBridge effectively strengthens feature discriminability and stabilizes decision boundaries under open-set conditions.

Experiments on the PalmVein dataset further verify the generalization capability of the proposed framework. Compared with the naive baseline, PalmBridge significantly improves CCNet, increasing ACC by 0.07\% and reducing EER by 0.2567\%, outperforming alternative frameworks and demonstrating strong adaptability to palm vein imagery. Other backbone networks also exhibit clear performance gains when integrated with PalmBridge, particularly in terms of EER reduction. Notably, the ACCs of CCNet and CO3Net even reach 100\%, which highlights the effectiveness of PalmBridge in challenging intra-dataset open-set scenarios.

\subsection{Cross-Dataset Open-Set Experiments}
To further evaluate the performance of the proposed PalmBridge, we conduct cross-dataset open-set experiments using CCNet as the backbone, with the results summarized in Tab.~\ref{tab:Cross}. In this protocol, one dataset is used for training, while the remaining datasets are used exclusively for testing. Compared with the naive baseline, PalmBridge consistently achieves lower EER in most cases, indicating improved robustness to distribution shifts across datasets. For example, when PalmBridge is trained on IITD and tested on PolyU and Tongji, the EER is reduced from 2.7646\% to 2.3435\% and from 3.1000\% to 2.8791\%, respectively. These results demonstrate that PalmBridge effectively mitigates cross-dataset feature discrepancies and improves verification performance on unseen data. Overall, the cross-dataset results confirm that PalmBridge not only enhances recognition accuracy within individual datasets but also substantially strengthens generalization across heterogeneous datasets. The consistent EER reductions observed under cross-dataset evaluation highlight the robustness of the proposed framework in handling acquisition-induced heterogeneity and unseen identity variations, which is critical for real-world deployment scenarios where training and testing data often originate from different environments

\subsection{Closed-Set Experiments}
Although our method is primarily designed for open-set recognition, we also evaluate its performance under closed-set scenarios to provide a comprehensive evaluation. In this experiment, CCNet is adopted as the backbone, and the results are reported in Tab. \ref{tab:Close}. PalmBridge attains an EER of 0.00\% on PolyU, which outperforms traditional coding-based methods such as PalmCode (0.35000\%) and Fusion Code (0.24000\%), as well as recent DL-based approaches, including CompNet~ (0.05560\%), CO3Net~ (0.02200\%), and CCNet (0.00006\%).
Consistent improvements are also observed on Tongji and IITD, where PalmBridge achieves EERs of 0.0083\% and 0.36\%, respectively, which are significantly lower than those of many existing approaches. These findings suggest that PalmBridge enhances feature discriminability even in closed scenarios, enabling clearer separation between identities.

A similar trend is observed on the PalmVein dataset, with results summarized in Tab. \ref{tab:CloseNIR}. PalmBridge achieves an EER of 0.00069\%, outperforming a variety of handcrafted feature-based and DL-based approaches. The consistently low EER values confirm the strong discriminability of PalmBridge and highlight its potential across different acquisition modalities and verification scenarios.

\begin{figure}[!t]
\centerline{\includegraphics[width=.85\columnwidth]{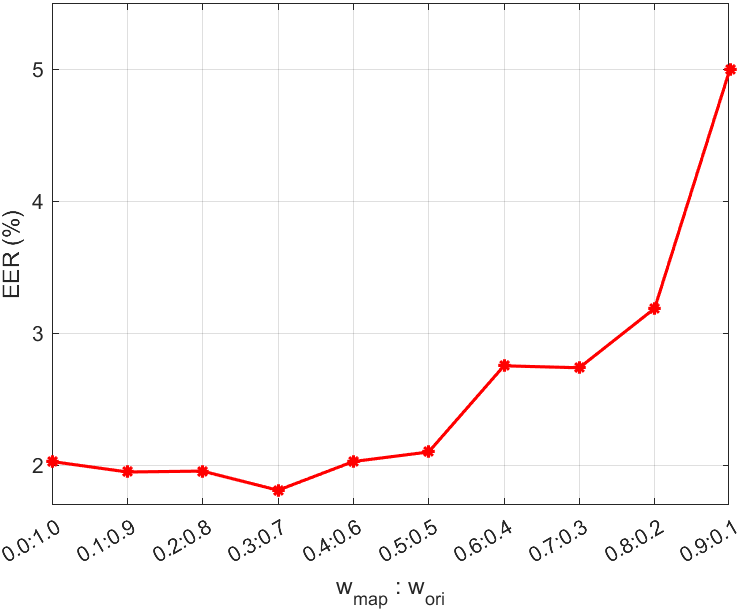}}
\caption{EERs under different blending coefficients.}
\label{fig:weights}
\end{figure}

\begin{table}[!t]
\centering
\caption{EERs(\%) of Ablation Results for CompNet.}
\label{tab:ablaComp}
\begin{tabular}{ccccccc}
\hline
\centering
 $\mathcal{L}_{\mathrm{bak}}$  & $\mathcal{L}_{\mathrm{con}}$  & $\mathcal{L}_{\mathrm{o}}$ & IITD & PolyU & Tongji & PalmVein 
                \\ \hline
\ding{51} & \ding{55} & \ding{55} &2.0290 &1.0741 &0.1733 &0.4111 \\ 
\ding{51} & \ding{51} & \ding{55} &1.8116 &0.9607 &0.1600 &\textbf{0.2771} \\   
\ding{51} & \ding{51} & \ding{51} &\textbf{1.3534} &\textbf{0.8254} &\textbf{0.1333} &0.2778  \\ \hline
\end{tabular}
\end{table}

\begin{table}[!t]
\centering
\caption{EERs(\%) of Ablation Results for CCNet.}
\label{tab:ablaCC}
\begin{tabular}{ccccccc}
\hline
\centering
 $\mathcal{L}_{\mathrm{bak}}$  & $\mathcal{L}_{\mathrm{con}}$  & $\mathcal{L}_{\mathrm{o}}$ & IITD & PolyU & Tongji & PalmVein 
                \\ \hline
\ding{51} & \ding{55} & \ding{55} &2.2679 &1.7831 &0.2131 &0.5577\\ 
\ding{51} & \ding{51} & \ding{55} &2.0695 &1.3289 &\textbf{0.1333} &0.3347 \\   
\ding{51} & \ding{51} & \ding{51} &\textbf{2.0689} &\textbf{1.2222} &\textbf{0.1333} &\textbf{0.3020}\\ \hline
\end{tabular}
\end{table}

\subsection{ROC Curves and GI Curves}
In addition to the quantitative results presented above, we further analyze the performance of PalmBridge using ROC and GI curves. Fig.~\ref{fig:roc} shows the ROC curves of CompNet and CCNet on four datasets under the naive and PalmBridge frameworks (solid: naive; dashed: PalmBridge). Across all datasets, PalmBridge consistently increases the GAR at the same FAR, including in the low-FAR region. This behavior indicates improved robustness that does not come from relaxing the decision boundary, as collapse-driven gains would typically lead to degraded performance at low FARs.

Fig.~\ref{fig:GI} provides further insight by comparing the GI curves of CCNet on PolyU and PalmVein. With PalmBridge, the Genuine distribution becomes more concentrated and shifts toward smaller distances, while the Imposter distribution exhibits no comparable left shift and only limited changes in shape. Consequently, the overlap between the genuine and impostor distributions is significantly reduced. These observations suggest that the performance gains of PalmBridge stem primarily from suppressing within-identity variation under heterogeneity, while preserving the relative separation between different identities.

\begin{table}[!t]
\centering
\caption{EERs(\%) of Ablation Results for CO3Net.}
\label{tab:ablaCO3}
\begin{tabular}{ccccccc}
\hline
\centering
 $\mathcal{L}_{\mathrm{bak}}$  & $\mathcal{L}_{\mathrm{con}}$  & $\mathcal{L}_{\mathrm{o}}$ & IITD & PolyU & Tongji & PalmVein 
                \\ \hline
\ding{51} & \ding{55} & \ding{55} &2.6812 &1.8385 &0.2000 &0.4481\\ 
\ding{51} & \ding{51} & \ding{55} &2.3755 &1.4444 &0.1867 &0.4000 \\   
\ding{51} & \ding{51} & \ding{51} &\textbf{2.1739} &\textbf{1.4232} &\textbf{0.1733} &\textbf{0.3667}\\ \hline
\end{tabular}
\end{table}

\begin{table}[!t]
\centering
\caption{EERs(\%) of Ablation Results for ResNet18.}
\label{tab:ablaRes18}
\begin{tabular}{ccccccc}
\hline
\centering
 $\mathcal{L}_{\mathrm{bak}}$  & $\mathcal{L}_{\mathrm{con}}$  & $\mathcal{L}_{\mathrm{o}}$ & IITD & PolyU & Tongji & PalmVein 
                \\ \hline
\ding{51} & \ding{55} & \ding{55} &7.3188 &9.5714 &5.2638 &9.3778\\ 
\ding{51} & \ding{51} & \ding{55} &6.1594 &\textbf{8.9683} &4.8394 &8.2778 \\   
\ding{51} & \ding{51} & \ding{51} &\textbf{5.8942} &9.4646 &\textbf{4.8177}&\textbf{8.2525}\\ \hline
\end{tabular}
\end{table}

\begin{table}[!t]
\centering
\caption{EERs(\%) of Ablation Results for DHN.}
\label{tab:ablaRes50}
\begin{tabular}{ccccccc}
\hline
\centering
 $\mathcal{L}_{\mathrm{bak}}$  & $\mathcal{L}_{\mathrm{con}}$  & $\mathcal{L}_{\mathrm{o}}$ & IITD & PolyU & Tongji & PalmVein 
                \\ \hline
\ding{51} & \ding{55} & \ding{55} &6.5217 &9.0488 &5.9554 &11.9667\\ 
\ding{51} & \ding{51} & \ding{55} &6.2765 &9.5132  &5.8172 &\textbf{8.5982} \\  
\ding{51} & \ding{51} & \ding{51} &\textbf{6.0870} &\textbf{8.9101} &\textbf{5.7200} &9.0284\\ \hline
\end{tabular}
\end{table}

\begin{table}[!t]
\centering
\caption{EERs(\%) of Ablation Results for SF2Net.}
\label{tab:ablaSF}
\begin{tabular}{ccccccc}
\hline
\centering
 $\mathcal{L}_{\mathrm{bak}}$  & $\mathcal{L}_{\mathrm{con}}$  & $\mathcal{L}_{\mathrm{o}}$ & IITD & PolyU & Tongji & PalmVein 
                \\ \hline
\ding{51} & \ding{55} & \ding{55} &2.3188 &1.4921 &0.1867 &0.3444\\ 
\ding{51} & \ding{51} & \ding{55} &2.2040 &\textbf{1.4180} &\textbf{0.1733} &0.2946 \\  
\ding{51} & \ding{51} & \ding{51} &\textbf{1.7391} &1.4550 &\textbf{0.1733} &\textbf{0.2889}\\ \hline
\end{tabular}
\end{table}

\begin{table}[!t]
\centering
\caption{EERs(\%) of Ablation Results with Different Numbers of PalmBridge Vectors.}
\begin{tabular}{cccc} %
\hline
Number of vectors & IITD & PolyU & Tongji \\ \hline
768 & 2.4638 & 1.6271 & 0.2667 
 \\ \hline
512 & 2.0689 & 1.2222 & 0.1333 
 \\ \hline
256 & 2.4885 & 1.5276 & 0.2301
 \\ \hline
128 & 2.3188 & 1.5276 & 0.1733 
 \\ \hline
64  & 2.3913 & 1.7885 & 0.2533 
 \\ \hline
\end{tabular}
\label{tab:numberabla}
\end{table}

\begin{table*}[!t]
\centering
\caption{Ablation Results of the plug-and-play experiments.}
\begin{tabular}{llcccccccc}
\hline
\multirow{2}{*}{Method}               & \multirow{2}{*}{Framework}   & \multicolumn{2}{c}{IITD} & \multicolumn{2}{c}{PolyU} & \multicolumn{2}{c}{Tongji} & \multicolumn{2}{c}{PalmVein}\\ \cmidrule(lr){3-4} \cmidrule(lr){5-6} \cmidrule(lr){7-8} \cmidrule(lr){9-10}
             &     & ACC(\%)$\uparrow$        & EER(\%)$\downarrow$       & ACC(\%)$\uparrow$           & EER(\%)$\downarrow$         & ACC(\%)$\uparrow$           & EER(\%)$\downarrow$    & ACC(\%)$\uparrow$           & EER(\%)$\downarrow$         \\ \hline

\multirow{2}{*}{CompNet \cite{liang2021compnet}} & Naive &            98.26&           2.0290&             98.84&            1.0741&             100.00&            0.1733&     99.87    &  0.4111          \\

&\cellcolor{lightergray}\textbf{PalmBridge} &\cellcolor{lightergray}\textbf{98.48}&\cellcolor{lightergray}\textbf{1.8116}&\cellcolor{lightergray}\textbf{99.21}&\cellcolor{lightergray}\textbf{1.0317} &\cellcolor{lightergray}\textbf{100.00}       &\cellcolor{lightergray}\textbf{0.1600}  &\cellcolor{lightergray}\textbf{100.00}       &\cellcolor{lightergray}\textbf{0.2752} \\
 \hline

\multirow{2}{*}{CO3Net \cite{yang2023co}} & Naive &             97.82&           2.6812&              98.10&            1.8385&             100.00&           0.2000&   99.93&    0.4481           \\

&\cellcolor{lightergray}\textbf{PalmBridge} &\cellcolor{lightergray}\textbf{98.04}&\cellcolor{lightergray}\textbf{2.6812}&\cellcolor{lightergray}\textbf{98.25}&\cellcolor{lightergray}\textbf{1.8466}&\cellcolor{lightergray}\textbf{100.00}       &\cellcolor{lightergray}\textbf{0.2000}  &\cellcolor{lightergray}\textbf{100.00}       &\cellcolor{lightergray}\textbf{0.3222} \\ \hline

\multirow{2}{*}{CCNet \cite{yang2023comprehensive}} & Naive &             97.39&           2.2679&              98.57&            1.7831&             100.00&           0.2131&    99.93&    0.5577           \\

&\cellcolor{lightergray}\textbf{PalmBridge} &\cellcolor{lightergray}\textbf{97.83}&\cellcolor{lightergray}\textbf{2.2464}&\cellcolor{lightergray}\textbf{98.36}&\cellcolor{lightergray}\textbf{1.7355}&\cellcolor{lightergray}\textbf{100.00}       &\cellcolor{lightergray}\textbf{0.2068}  &\cellcolor{lightergray}\textbf{100.00}       &\cellcolor{lightergray}\textbf{0.4842}\\ \hline
\end{tabular}
\label{tab:AblaPlug}
\end{table*}

\begin{table*}[!t]
\centering
\caption{Time cost comparison of different components in PalmBridge.}
\begin{tabular}{lcccc} %
\hline
Method & Backbone(s) & PalmBridge Module(s) & Total(s) & Module Overhead(\%)\\ \hline

ResNet18\cite{he2016deep} & 3.6682 & 0.2304 & 3.8986 & 5.91
 \\ \hline

DHN\cite{wu2021palmprint} & 4.0438 & 0.1415 & 4.1853 & 3.38
 \\ \hline

CompNet\cite{liang2021compnet} & 3.4340 & 0.4180 & 3.8520 & 10.85
 \\ \hline

CCNet\cite{yang2023comprehensive} & 4.1021 & 0.1821 & 4.2842 & 4.25
 \\ \hline

CO3Net\cite{yang2023co} & 5.7094 & 0.1609 & 5.8703 & 2.74
 \\ \hline

SF2Net\cite{liu2025sf2net} & 3.8284 & 0.1609 & 3.9893 & 4.03
 \\ \hline
\end{tabular}
\label{tab:timecost}
\end{table*}

\subsection{Ablation Study}
\subsubsection{Blending coefficients}
To analyze the effect of different blending coefficients in Eq.~\eqref{eq:weighted mapped}, we conduct an ablation study on IITD using CompNet as the backbone. The visualization results are presented in Fig.~\ref{fig:weights}. It is evident that the EER exhibits a non-linear relationship with respect to the blending ratio, decreasing initially and then rising as the contribution of the mapping coefficient becomes dominant. Specifically, when the ratio of $w_{map}$: $w_{ori}$ increases from 0: 1 to approximately 0.3: 0.7, the EER remains relatively stable at around 2\%. Once $w_{map}$ exceeds 0.3, the EER begins to increase noticeably, reaching 5\% when the ratio is set to 0.9: 0.1.
This trend suggests that excessive reliance on mapped features may compromise feature integrity and degrade discriminative information. The optimal performance is achieved at $w_{map}$: $w_{ori}$ = 0.3:0.7, indicating that an appropriate balance between mapped and original features is crucial for robust verification. We infer that while feature mapping improves inter-domain alignment and suppresses nuisance variation, overly aggressive mapping can introduce information distortion, which causes multiple identities to converge toward the same learned vector. Retaining part of the original representation helps preserve fine-grained identity cues that cannot be fully captured by a finite set of learned vectors.

\subsubsection{Loss functions}
The impact of different loss components in Eq.~\eqref{eq:total loss} is evaluated across four datasets and multiple backbones, with the results presented in Tabs. \ref{tab:ablaComp}-\ref{tab:ablaSF}. 
When only the feature consistency loss is enabled, most models achieve a noticeable reduction in EER across all datasets, indicating that enforcing alignment between original and mapped representations enhances feature stability and robustness. When the orthogonality loss is further introduced alongside the consistency loss, the EER decreases even more. This observation suggests a complementary effect between the two loss terms, where consistency promotes reliable feature alignment between the backbone and PalmBridge, and the orthogonality loss encourages vector diversity. Together, these losses facilitate the learning of more discriminative and well-structured feature representations, leading to improved verification performance.

\subsubsection{Number of vectors}
To investigate the impact of the number of PalmBridge vectors on the overall framework, we conduct an ablation study using CCNet as the backbone across multiple datasets with varying vector counts. The results are summarized in Tab. \ref{tab:numberabla}. As shown, varying the number of vectors leads to only marginal changes in EER across the IITD, PolyU, and Tongji datasets. This observation indicates that PalmBridge is relatively insensitive to the vector cardinality. The framework remains robust across a wide range of vector numbers and does not rely on a carefully tuned configuration, which provides strong stability and practicality for real-world deployment.

\subsubsection{Plug-and-Play}
To validate the plug-and-play capability of the proposed PalmBridge framework, we further conduct an ablation study. Specifically, we first train CompNet, CO3Net, and CCNet under both the naive and PalmBridge frameworks using the intra-dataset protocol described in Sec. IV-A. As a result, for each dataset, we obtain three naive backbone models, three PalmBridge backbone models, and three corresponding PalmBridge vector sets.
During the inference phase, we incorporate the learned PalmBridge vector set $\mathbf{P}$ into the naive backbone model in a plug-and-play manner. Notably, neither the PalmBridge vector set nor the naive backbone networks are further tuned.
The comparative results are summarized in Tab \ref{tab:AblaPlug}. Despite the absence of joint optimization between the PalmBridge vectors and the backbone models, consistent performance improvements are still observed across datasets and architectures. This clearly demonstrates the plug-and-play nature of PalmBridge and its ability to enhance verification performance without requiring additional training or modification of the backbone networks.

\subsection{Time Cost Experiments}
To evaluate the impact of PalmBridge on inference efficiency, we measure the computational time of each processing component for six backbone networks integrated with PalmBridge on the IITD dataset. As shown in Tab. \ref{tab:timecost}, PalmBridge introduces a modest runtime overhead that varies across backbone architectures. On IITD using an RTX 3090, PalmBridge accounts for 2.74\%–10.85\% of the total inference time across the six backbones. These results indicate that the computational cost of PalmBridge is limited and controllable. In practice, runtime can be flexibly balanced against performance gains by adjusting the number of representative vectors and selecting an appropriate nearest-vector search implementation, allowing PalmBridge to be adapted to different deployment scenarios with varying latency constraints.


\section{Conclusion}
In this paper, we investigate open-set palmprint verification under feature-distribution shift. We propose PalmBridge, a lightweight feature-mapping module that learns a compact set of representative vectors and maps both gallery and query vectors into a shared embedding space during enrollment and verification. The representative vectors are jointly optimized using the proposed loss to promote diversity and reduce redundancy.
At inference time, PalmBridge assigns each feature to its nearest representative vector and blends the mapped representation with the original feature vector, effectively suppressing nuisance variation while preserving identity-discriminative information. We further provide a theoretical analysis that explains when genuine-pair dispersion contracts through consistent assignments and clarifies the risk of identity mixing caused by assignment collisions.
Extensive experiments across datasets, backbones, and evaluation protocols show consistent reductions in EER under both intra- and cross-dataset scenarios. Additional plug-and-play experiments demonstrate that PalmBridge can improve verification performance without retraining the backbone, while runtime profiling indicates that the introduced computational overhead remains modest. An interesting direction for future research is to extend PalmBridge to support cross-model feature mapping, which enables interoperability across heterogeneous backbone architectures.

\bibliographystyle{IEEEtran}
\bibliography{reference}

\begin{thebibliography}{10}
\providecommand{\url}[1]{#1}
\csname url@samestyle\endcsname
\providecommand{\newblock}{\relax}
\providecommand{\bibinfo}[2]{#2}
\providecommand{\BIBentrySTDinterwordspacing}{\spaceskip=0pt\relax}
\providecommand{\BIBentryALTinterwordstretchfactor}{4}
\providecommand{\BIBentryALTinterwordspacing}{\spaceskip=\fontdimen2\font plus
\BIBentryALTinterwordstretchfactor\fontdimen3\font minus \fontdimen4\font\relax}
\providecommand{\BIBforeignlanguage}[2]{{%
\expandafter\ifx\csname l@#1\endcsname\relax
\typeout{** WARNING: IEEEtran.bst: No hyphenation pattern has been}%
\typeout{** loaded for the language `#1'. Using the pattern for}%
\typeout{** the default language instead.}%
\else
\language=\csname l@#1\endcsname
\fi
#2}}
\providecommand{\BIBdecl}{\relax}
\BIBdecl

\bibitem{zhang2003online}
D.~Zhang, W.-K. Kong, J.~You, and M.~Wong, ``Online palmprint identification,'' \emph{IEEE Transactions on pattern analysis and machine intelligence}, vol.~25, no.~9, pp. 1041--1050, 2003.

\bibitem{gao2025deep}
C.~Gao, Z.~Yang, W.~Jia \emph{et~al.}, ``Deep learning in palmprint recognition: A comprehensive survey,'' \emph{IEEE Transactions on Systems, Man, and Cybernetics: Systems}, 2026.

\bibitem{zhou2022domain}
K.~Zhou, Z.~Liu, Y.~Qiao, T.~Xiang, and C.~C. Loy, ``Domain generalization: A survey,'' \emph{IEEE transactions on pattern analysis and machine intelligence}, vol.~45, no.~4, pp. 4396--4415, 2022.

\bibitem{shen2022distribution}
L.~Shen, Y.~Zhang, K.~Zhao, R.~Zhang, and W.~Shen, ``Distribution alignment for cross-device palmprint recognition,'' \emph{Pattern Recognition}, vol. 132, p. 108942, 2022.

\bibitem{deng2019arcface}
J.~Deng, J.~Guo, N.~Xue, and S.~Zafeiriou, ``Arcface: Additive angular margin loss for deep face recognition,'' in \emph{Proceedings of the IEEE/CVF conference on computer vision and pattern recognition}, 2019, pp. 4690--4699.

\bibitem{shao2024learning}
H.~Shao, Y.~Zou, C.~Liu \emph{et~al.}, ``Learning to generalize unseen dataset for cross-dataset palmprint recognition,'' \emph{IEEE Transactions on Information Forensics and Security}, vol.~19, pp. 3788--3799, 2024.

\bibitem{jia2025single}
C.~Jia, X.~Dong, Y.~L. Lai, A.~B.~J. Teoh, Z.~Yang, X.~Zhang, L.~Wang, Z.~Jin, and L.~Yang, ``Single source domain generalization for palm biometrics,'' \emph{Pattern Recognition}, vol. 165, p. 111620, 2025.

\bibitem{van2017neural}
A.~Van Den~Oord, O.~Vinyals \emph{et~al.}, ``Neural discrete representation learning,'' \emph{Advances in neural information processing systems}, vol.~30, 2017.

\bibitem{zhong2019decade}
D.~Zhong, X.~Du, and K.~Zhong, ``Decade progress of palmprint recognition: A brief survey,'' \emph{Neurocomputing}, vol. 328, pp. 16--28, 2019.

\bibitem{zhang2018combining}
S.~Zhang, H.~Wang, W.~Huang, and C.~Zhang, ``Combining modified lbp and weighted src for palmprint recognition,'' \emph{Signal, Image and Video Processing}, vol.~12, no.~6, pp. 1035--1042, 2018.

\bibitem{fei2021jointly}
L.~Fei, B.~Zhang, Y.~Xu \emph{et~al.}, ``Jointly heterogeneous palmprint discriminant feature learning,'' \emph{IEEE Transactions on Neural Networks and Learning Systems}, vol.~33, no.~9, pp. 4979--4990, 2021.

\bibitem{kong2004competitive}
A.-K. Kong and D.~Zhang, ``Competitive coding scheme for palmprint verification,'' in \emph{Proceedings of the International Conference on Pattern Recognition}, vol.~1.\hskip 1em plus 0.5em minus 0.4em\relax IEEE, 2004, pp. 520--523.

\bibitem{guo2009palmprint}
Z.~Guo, D.~Zhang, L.~Zhang, and W.~Zuo, ``Palmprint verification using binary orientation co-occurrence vector,'' \emph{Pattern Recognition Letters}, vol.~30, no.~13, pp. 1219--1227, 2009.

\bibitem{yang2023multi}
Z.~Yang, L.~Leng, T.~Wu, M.~Li, and J.~Chu, ``Multi-order texture features for palmprint recognition,'' \emph{Artificial Intelligence Review}, vol.~56, no.~2, pp. 995--1011, 2023.

\bibitem{chai2019boosting}
T.~Chai, S.~Prasad, and S.~Wang, ``Boosting palmprint identification with gender information using deepnet,'' \emph{Future Generation Computer Systems}, vol.~99, pp. 41--53, 2019.

\bibitem{genovese2019palmnet}
A.~Genovese, V.~Piuri, K.~N. Plataniotis, and F.~Scotti, ``Palmnet: Gabor-pca convolutional networks for touchless palmprint recognition,'' \emph{IEEE Transactions on Information Forensics and Security}, vol.~14, no.~12, pp. 3160--3174, 2019.

\bibitem{zhong2018palm}
D.~Zhong, S.~Liu, W.~Wang, and X.~Du, ``Palm vein recognition with deep hashing network,'' in \emph{Chinese Conference on Pattern Recognition and Computer Vision (PRCV)}.\hskip 1em plus 0.5em minus 0.4em\relax Springer, 2018, pp. 38--49.

\bibitem{zhu2016deep}
H.~Zhu, M.~Long, J.~Wang, and Y.~Cao, ``Deep hashing network for efficient similarity retrieval,'' in \emph{Proceedings of the AAAI conference on Artificial Intelligence}, vol.~30, no.~1, 2016.

\bibitem{yang2023comprehensive}
Z.~Yang, H.~Huangfu, L.~Leng \emph{et~al.}, ``Comprehensive competition mechanism in palmprint recognition,'' \emph{IEEE Transactions on Information Forensics and Security}, vol.~18, pp. 5160--5170, 2023.

\bibitem{liang2021compnet}
X.~Liang, J.~Yang, G.~Lu, and D.~Zhang, ``Compnet: Competitive neural network for palmprint recognition using learnable gabor kernels,'' \emph{IEEE Signal Processing Letters}, vol.~28, pp. 1739--1743, 2021.

\bibitem{yang2023co}
Z.~Yang, W.~Xia, Y.~Qiao \emph{et~al.}, ``Co3net: Coordinate-aware contrastive competitive neural network for palmprint recognition,'' \emph{IEEE Transactions on Instrumentation and Measurement}, vol.~72, pp. 1--14, 2023.

\bibitem{du2020cross}
X.~Du, D.~Zhong, and H.~Shao, ``Cross-domain palmprint recognition via regularized adversarial domain adaptive hashing,'' \emph{IEEE Transactions on Circuits and Systems for Video Technology}, vol.~31, no.~6, pp. 2372--2385, 2020.

\bibitem{shao2021towards}
H.~Shao and D.~Zhong, ``Towards cross-dataset palmprint recognition via joint pixel and feature alignment,'' \emph{IEEE Transactions on Image Processing}, vol.~30, pp. 3764--3777, 2021.

\bibitem{fei2023learning}
L.~Fei, W.~K. Wong, S.~Zhao \emph{et~al.}, ``Learning spectrum-invariance representation for cross-spectral palmprint recognition,'' \emph{IEEE Transactions on Systems, Man, and Cybernetics: Systems}, vol.~53, no.~6, pp. 3868--3879, 2023.

\bibitem{zhong2025regpalm}
Y.~Zhong, W.~Chai, L.~Wang \emph{et~al.}, ``Regpalm: Towards large-scale open-set palmprint recognition by reducing pattern variance,'' \emph{IEEE Transactions on Information Forensics and Security}, 2025.

\bibitem{he2016deep}
K.~He, X.~Zhang, S.~Ren, and J.~Sun, ``Deep residual learning for image recognition,'' in \emph{Proceedings of the IEEE conference on computer vision and pattern recognition}, 2016, pp. 770--778.

\bibitem{wu2021palmprint}
T.~Wu, L.~Leng, M.~K. Khan, and F.~A. Khan, ``Palmprint-palmvein fusion recognition based on deep hashing network,'' \emph{IEEE access}, vol.~9, pp. 135\,816--135\,827, 2021.

\bibitem{liu2025sf2net}
Y.~Liu, L.~Leng, Z.~Yang, A.~B.~J. Teoh, and B.~Zhang, ``Sf2net: Sequence feature fusion network for palmprint verification,'' \emph{IEEE Transactions on Information Forensics and Security}, 2025.

\bibitem{kong2006palmprint}
A.~Kong, D.~Zhang, and M.~Kamel, ``Palmprint identification using feature-level fusion,'' \emph{Pattern Recognition}, vol.~39, no.~3, pp. 478--487, 2006.

\bibitem{jia2008palmprint}
W.~Jia, D.-S. Huang, and D.~Zhang, ``Palmprint verification based on robust line orientation code,'' \emph{Pattern Recognition}, vol.~41, no.~5, pp. 1504--1513, 2008.

\bibitem{zhang2012fragile}
L.~Zhang, H.~Li, and J.~Niu, ``Fragile bits in palmprint recognition,'' \emph{IEEE Signal processing letters}, vol.~19, no.~10, pp. 663--666, 2012.

\bibitem{fei2016half}
L.~Fei, Y.~Xu, and D.~Zhang, ``Half-orientation extraction of palmprint features,'' \emph{Pattern Recognition Letters}, vol.~69, pp. 35--41, 2016.

\bibitem{xu2018discriminative}
Y.~Xu, L.~Fei, J.~Wen \emph{et~al.}, ``Discriminative and robust competitive code for palmprint recognition,'' \emph{IEEE Transactions on Systems, Man, and Cybernetics: Systems}, vol.~48, no.~2, pp. 232--241, 2018.

\bibitem{yang2020extreme}
Z.~Yang, L.~Leng, and W.~Min, ``Extreme downsampling and joint feature for coding-based palmprint recognition,'' \emph{IEEE Transactions on Instrumentation and Measurement}, vol.~70, pp. 1--12, 2020.

\bibitem{zhang2017towards}
L.~Zhang, L.~Li, A.~Yang, Y.~Shen, and M.~Yang, ``Towards contactless palmprint recognition: A novel device, a new benchmark, and a collaborative representation based identification approach,'' \emph{Pattern Recognition}, vol.~69, pp. 199--212, 2017.

\bibitem{kumar2010personal}
A.~Kumar and S.~Shekhar, ``Personal identification using multibiometrics rank-level fusion,'' \emph{IEEE Transactions on Systems, Man, and Cybernetics, Part C (Applications and Reviews)}, vol.~41, no.~5, pp. 743--752, 2010.

\bibitem{zhang2009online}
D.~Zhang, Z.~Guo, G.~Lu, L.~Zhang, and W.~Zuo, ``An online system of multispectral palmprint verification,'' \emph{IEEE transactions on instrumentation and measurement}, vol.~59, no.~2, pp. 480--490, 2009.

\bibitem{zhong2019centralized}
D.~Zhong and J.~Zhu, ``Centralized large margin cosine loss for open-set deep palmprint recognition,'' \emph{IEEE Transactions on Circuits and Systems for Video Technology}, vol.~30, no.~6, pp. 1559--1568, 2019.

\bibitem{jin2025unified}
J.~Jin, C.~Zhao, R.~Zhang, S.~Shang, Y.~Zhao, J.~Wang, J.~Zhang, S.~Ding, W.~Jia, and Y.~Wu, ``Unified adversarial augmentation for improving palmprint recognition,'' in \emph{Proceedings of the IEEE/CVF International Conference on Computer Vision}, 2025, pp. 14\,141--14\,151.

\end{thebibliography}

\begin{IEEEbiography}[{\includegraphics[width=1in,height=1.25in,clip,keepaspectratio]{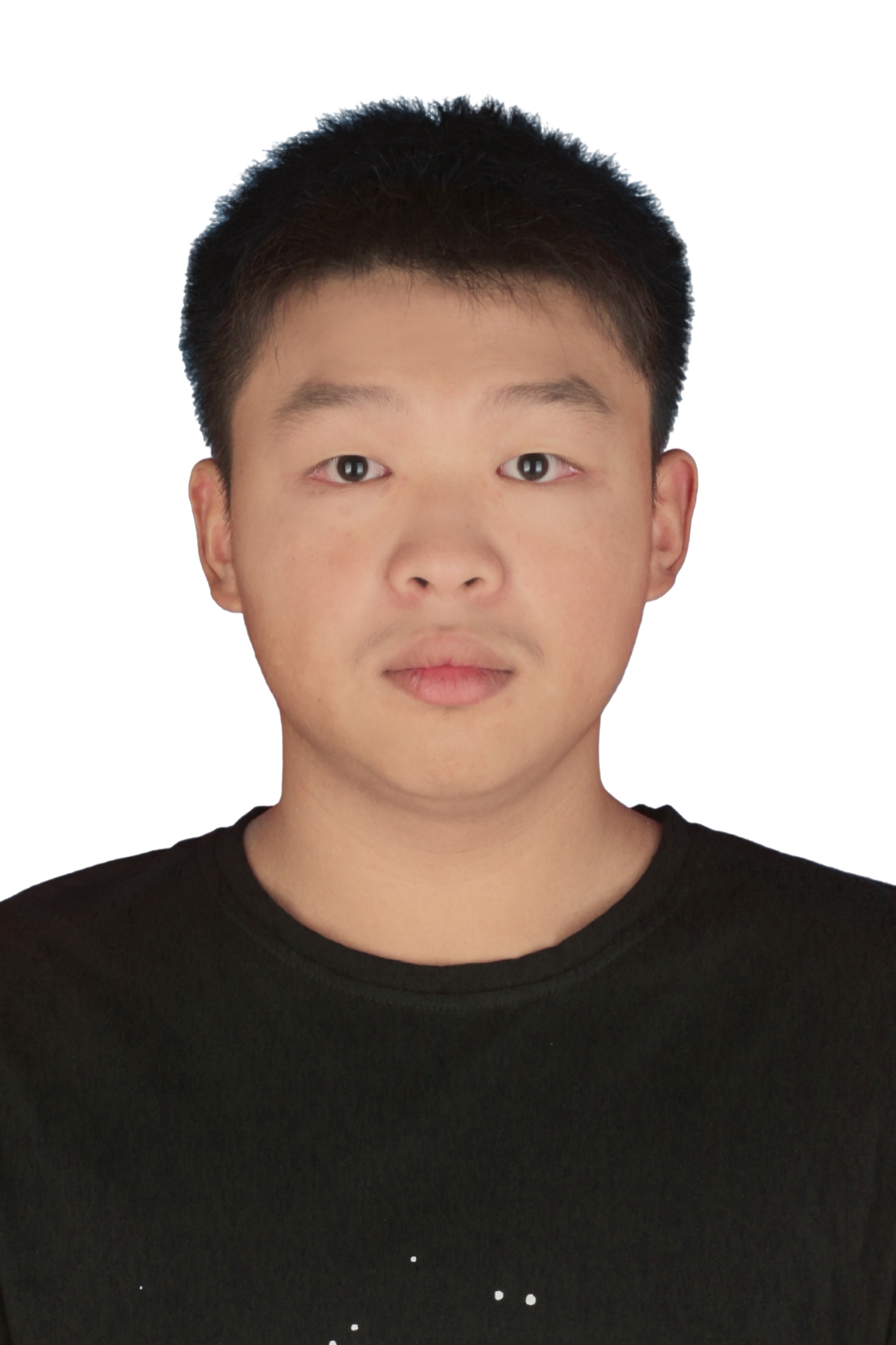}}]{Chenke Zhang}
received the B.S. degree in the School of Cyber Science and Engineering, Sichuan University, Chengdu, China, in 2025, where he is currently pursuing the M.S. degree. His research interests include biometrics.
\end{IEEEbiography}

\begin{IEEEbiography}[{\includegraphics[width=1in,height=1.25in,clip,keepaspectratio]{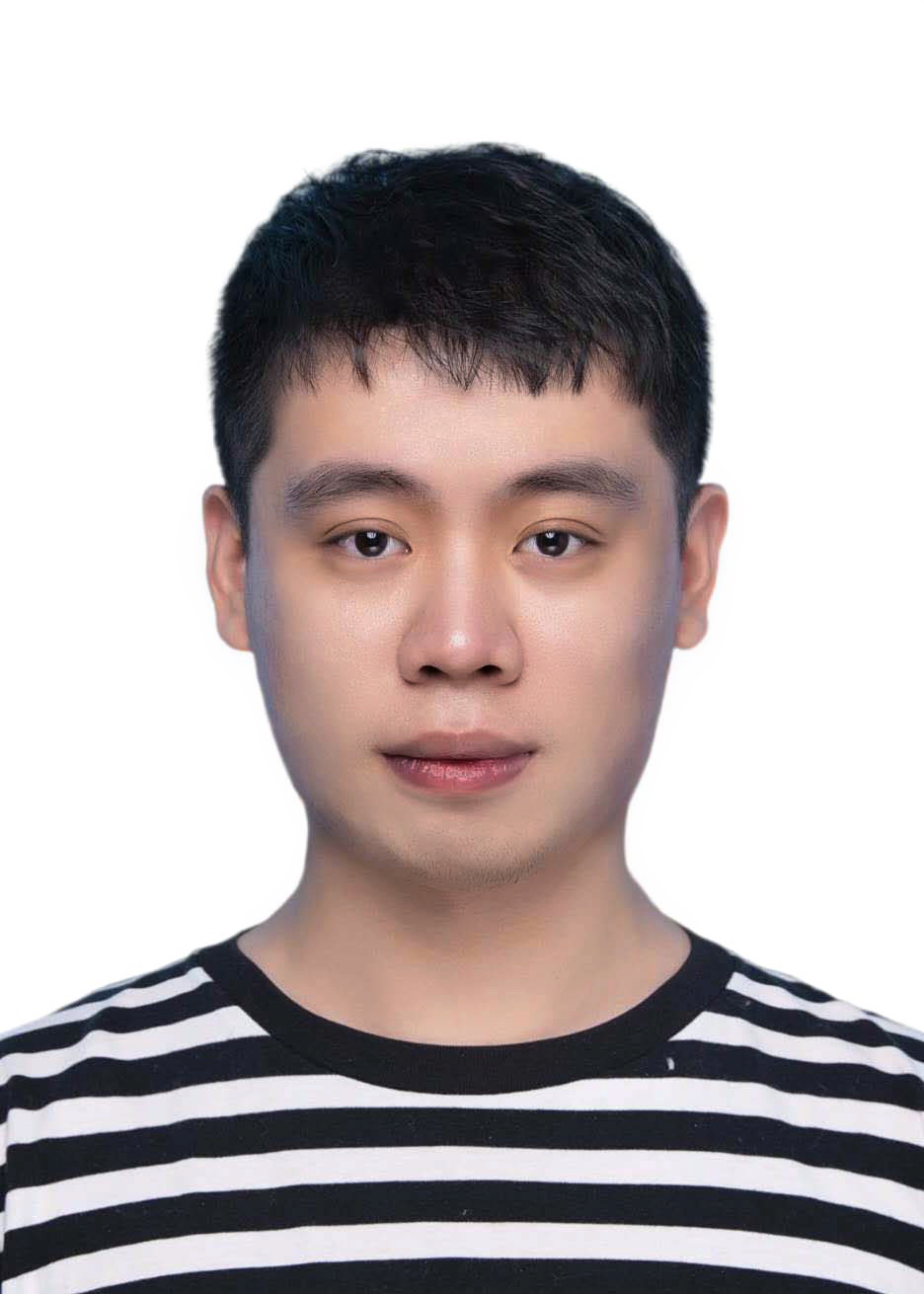}}]{Ziyuan Yang} is currently a researcher at Sichuan University. He received the M.S. degree in computer science from the School of Information Engineering, Nanchang University, Nanchang, China, in 2021, and the Ph.D. degree from the College of Computer Science, Sichuan University, China. He was a Research Intern at the Centre for Frontier AI Research, Agency for Science, Technology and Research (A*STAR), Singapore. In the last few years, he has published over 50 papers in leading machine learning conferences and journals, including CVPR, AAAI, IJCV, IEEE T-IFS, IEEE T-NNLS, IEEE T-SMCS, and IEEE T-AI. He was a reviewer for leading journals or conferences, e.g. IEEE T-PAMI, IEEE T-TIP, IEEE T-IFS, IEEE T-MI, CVPR, and ICCV.
\end{IEEEbiography}

\begin{IEEEbiography}[{\includegraphics[width=1in,height=1.25in,clip,keepaspectratio]{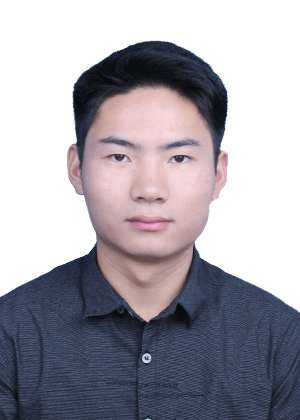}}]{Licheng Yan}
received the M.S. degree in Software Engineering from Nanchang Hangkong University, China in 2024. He is currently pursuing the Ph.D. degree in Computer Science at the PAMI Group, Faculty of Science and Technology, University of Macau. His research interests include biometrics, security analysis, and image generation. He was the reviewer for leading journals or conferences, including TIP, AAAI, and IJCB.
\end{IEEEbiography}

\begin{IEEEbiography}[{\includegraphics[width=1in,height=1.25in,clip,keepaspectratio]{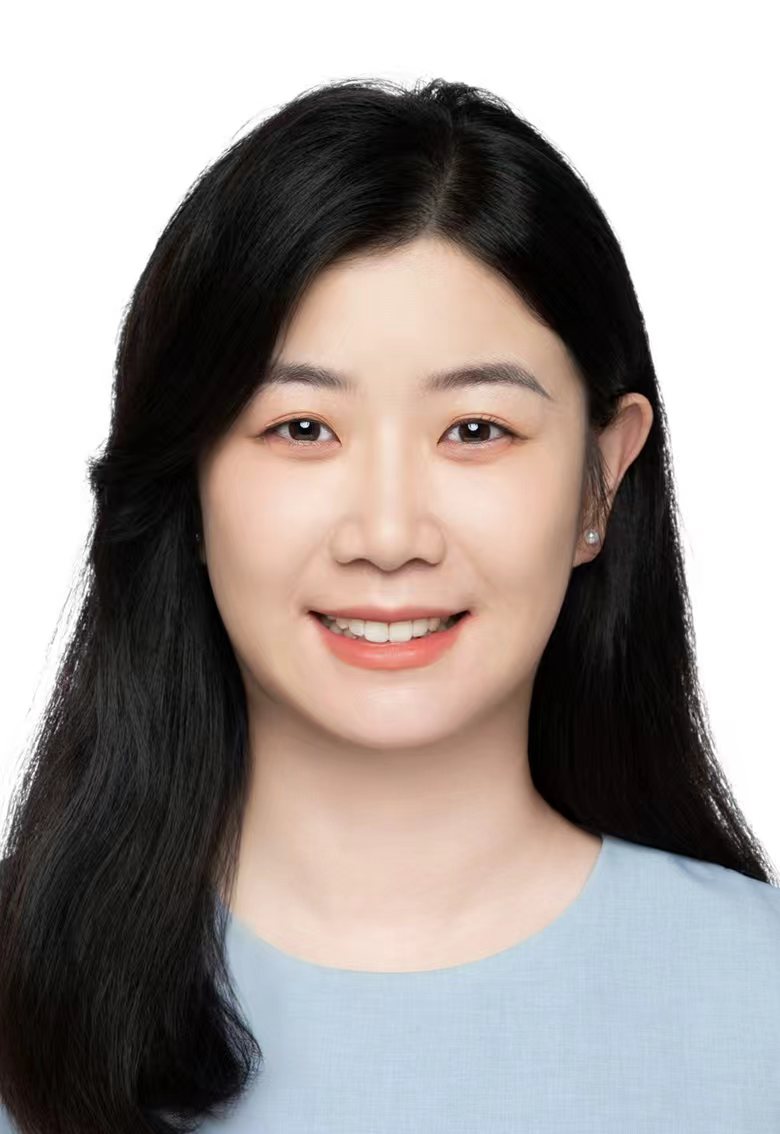}}]{Shuyi Li} received the Ph.D. degree in the Faculty of Science and Technology, University of Macau, Macau, China, in 2022. She is currently an Associate Researcher with the School of Information Science and Technology, Beijing University of Technology, Beijing, China. Her research interests include pattern recognition, multimodal biometrics, and multiview learning.
\end{IEEEbiography}

\begin{IEEEbiography}[{\includegraphics[width=1in,height=1.25in,clip,keepaspectratio]{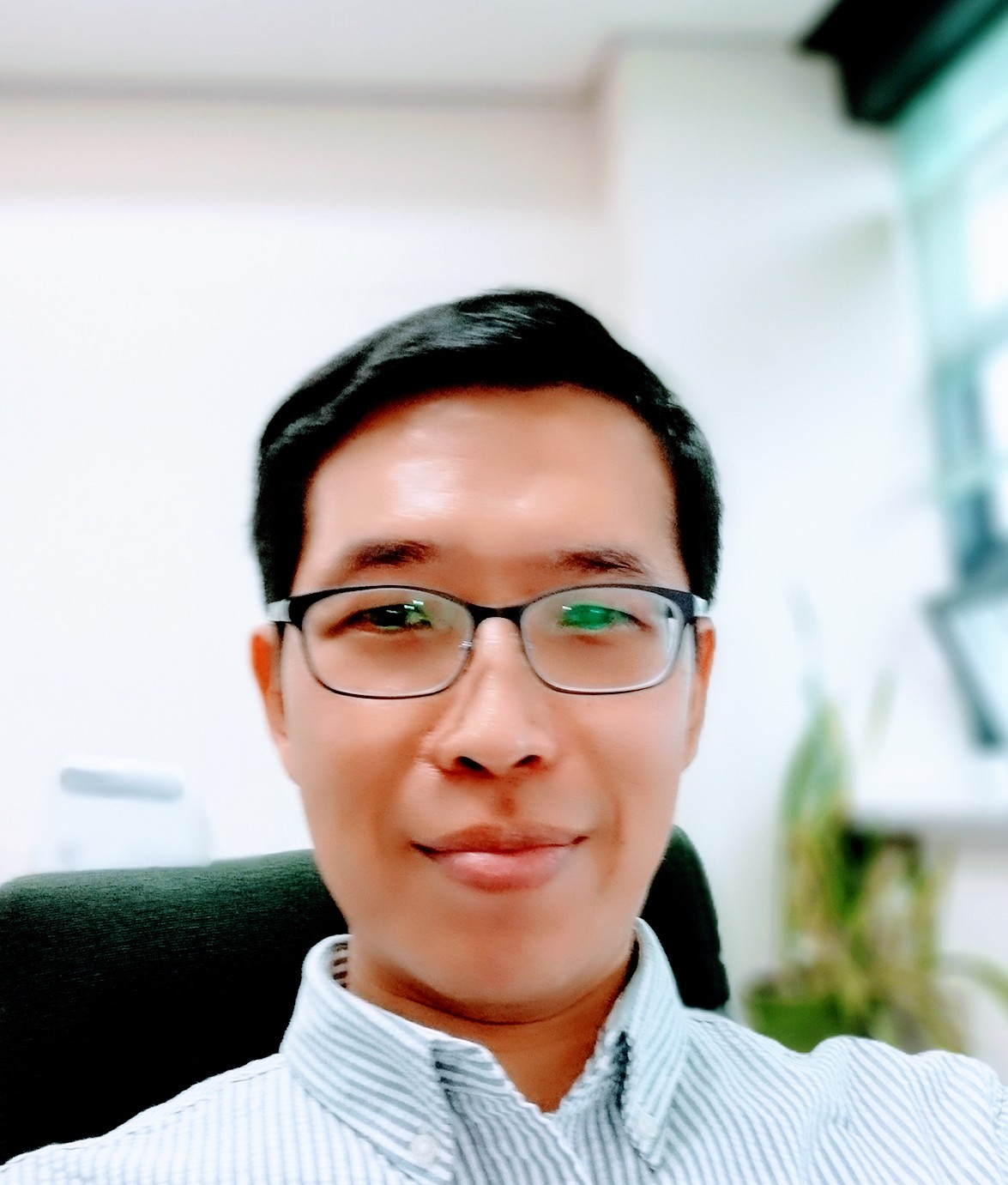}}]{Andrew Beng Jin Teoh}
(Senior Member, IEEE) received the B.Eng. degree in electronics and the Ph.D. degree from the National University of Malaysia in 1999 and 2003, respectively. He is currently a Full Professor with the Electrical and Electronic Engineering Department, College of Engineering, Yonsei University, South Korea. He has published over 350 international refereed journal articles and conference papers and edited several book chapters and volumes. His research, for which he has received funding, focuses on biometric applications and biometric security. His current research interests include machine learning and information security. He was the Guest Editor of the IEEE Signal Processing Magazine and a Senior Associate Editor of IEEE Transactions on Information Forensics and Security, IEEE Biometrics Compendium, and Machine Learning with Applications.
\end{IEEEbiography}

\begin{IEEEbiography}[{\includegraphics[width=1in,height=1.25in,clip,keepaspectratio]{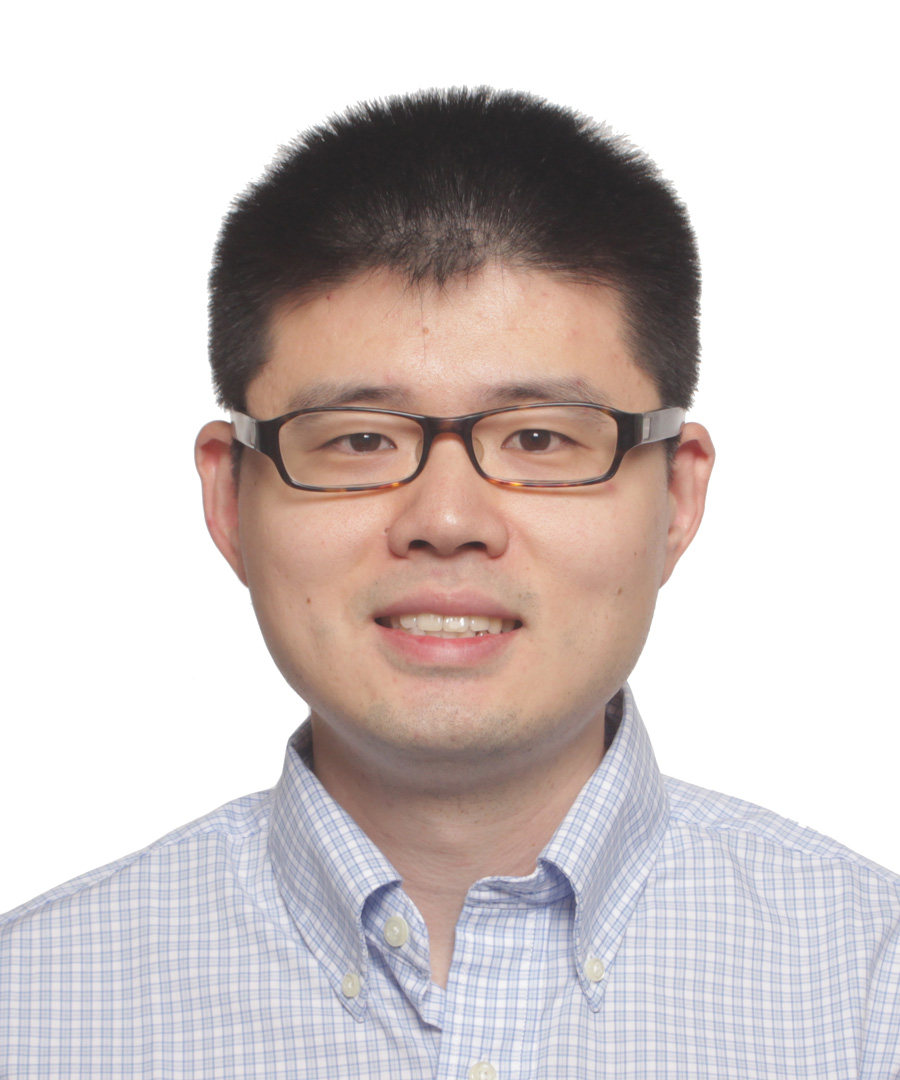}}]{Bob Zhang}
(Senior Member, IEEE) received the B.A. degree in computer science from York University, Toronto, ON, Canada, in 2006, the M.A.Sc. degree in information systems security from Concordia University, Montreal, QC, Canada, in 2007, and the Ph.D. in electrical and computer engineering from the University of Waterloo, Waterloo, ON, Canada, in 2011. After graduating from the University of Waterloo, he remained with the Center for Pattern Recognition and Machine Intelligence, and later, he was a Postdoctoral Researcher with the Department of Electrical and Computer Engineering, Carnegie Mellon University, Pittsburgh, PA, USA. He is currently an Associate Professor with the Department of Computer and Information Science, University of Macau, Macau. His research interests focus on biometrics, pattern recognition, and image processing. Dr. Zhang is a Technical Committee Member of the IEEE Systems, Man, and Cybernetics Society and an Associate Editor of IEEE TRANSACTIONS ON NEURAL NETWORKS AND LEARNING SYSTEMS, Artificial Intelligence Review, and IET Computer Vision.\end{IEEEbiography}

\begin{IEEEbiography}[{\includegraphics[width=1in,height=1.25in,clip,keepaspectratio]{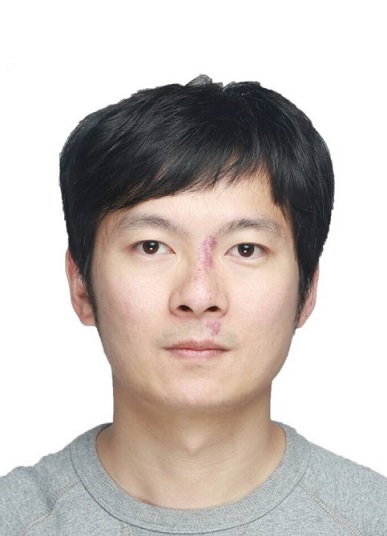}}]{Yi Zhang}
(Senior Member, IEEE) received the B.S., M.S., and Ph.D. degrees in computer science and technology from the College of Computer Science, Sichuan University, Chengdu, China, in 2005, 2008, and 2012, respectively. From 2014 to 2015, he was with the Department of Biomedical Engineering, Rensselaer Polytechnic Institute, Troy, NY, USA, as a Post-Doctoral Researcher. He is currently a Full Professor with the School of Cyber Science and Engineering, Sichuan University, and the Director of the Deep Imaging Group (DIG). He has authored more than 80 papers in the field of medical imaging, artificial intelligence, and information security. These papers were published in several leading journals, including IEEE Transactions on Medical Imaging, IEEE Transactions on Computational Imaging, Medical Image Analysis, European Radiology, and Optics Express, and reported by the Institute of Physics (IOP) and during the Lindau Nobel Laureate Meeting. He received major funding from the National Key Research and Development Program of China, the National Natural Science Foundation of China, and the Science and Technology Support Project of Sichuan Province, China. His research interests include medical imaging, compressive sensing, and deep learning. He is a Guest Editor of the International Journal of Biomedical Imaging and Sensing and Imaging and an Associate Editor of IEEE Transactions on Medical Imaging and IEEE Transactions on Radiation and Plasma Medical Sciences.
\end{IEEEbiography}

\end{document}